\pdfoutput=1

\documentclass[11pt]{article}

\usepackage[final]{coling}

\usepackage{times}
\usepackage{latexsym}

\usepackage[T1]{fontenc}

\usepackage[utf8]{inputenc}

\usepackage{microtype}

\usepackage{inconsolata}

\usepackage{graphicx}
\usepackage{csquotes}
\usepackage{enumitem}
\usepackage{amsmath}
\usepackage{algorithm}
\usepackage{bbding}
\usepackage{algpseudocode}
\usepackage{placeins}

\usepackage{tabularx}
\usepackage{booktabs}
\usepackage{array}
\usepackage{multirow}
\usepackage{xcolor}
\usepackage{caption}
\usepackage{cleveref}
\usepackage{makecell}
\usepackage{pdflscape}
\usepackage{geometry}
\usepackage{longtable}
\usepackage{hyperref,stackengine}

\definecolor{ForestGreen}{RGB}{34,139,34}
\newcolumntype{M}[1]{>{\centering\arraybackslash}m{#1}}
\newcolumntype{L}{>{\raggedright\arraybackslash}X}
\newcolumntype{C}[1]{>{\centering\arraybackslash}p{#1}}
\newcolumntype{M}[1]{>{\raggedright\arraybackslash}p{#1}}

\title{GroUSE: A Benchmark to Evaluate Evaluators in Grounded Question Answering}

\author{Sacha Muller \quad António Loison\thanks{For inquiries, please contact these authors.} \quad Bilel Omrani \quad Gautier Viaud\footnotemark[1] \\ Illuin Technology
\\
\small\texttt{\{ sacha.muller, antonio.loison, bilel.omrani, gautier.viaud \}@illuin.tech}}

\begin{document}
\maketitle
\begin{abstract}
Retrieval-Augmented Generation (RAG) has emerged as a common paradigm to use Large Language Models (LLMs) alongside private and up-to-date knowledge bases. In this work, we address the challenges of using LLM-as-a-Judge when evaluating grounded answers generated by RAG systems. To assess the calibration and discrimination capabilities of judge models, we identify 7 generator failure modes and introduce GroUSE \textit{(Grounded QA Unitary Scoring of Evaluators)}, a meta-evaluation benchmark of 144 unit tests. This benchmark reveals that existing automated RAG evaluation frameworks often overlook important failure modes, even when using \mbox{GPT-4} as a judge.

To improve on the current design of automated RAG evaluation frameworks, we propose a novel pipeline and find that while closed models perform well on GroUSE, state-of-the-art open-source judges do not generalize to our proposed criteria, despite strong correlation with GPT-4's judgement. Our findings suggest that correlation with GPT-4 is an incomplete proxy for the practical performance of judge models and should be supplemented with evaluations on unit tests for precise failure mode detection.

We further show that finetuning \mbox{Llama-3} on GPT-4's reasoning traces significantly boosts its evaluation capabilities, improving upon both correlation with GPT-4's evaluations and calibration on reference situations.  \footnote{\url{https://github.com/illuin-tech/grouse}}

\end{abstract}

\begin{figure*}
  \includegraphics[width=0.96\linewidth]{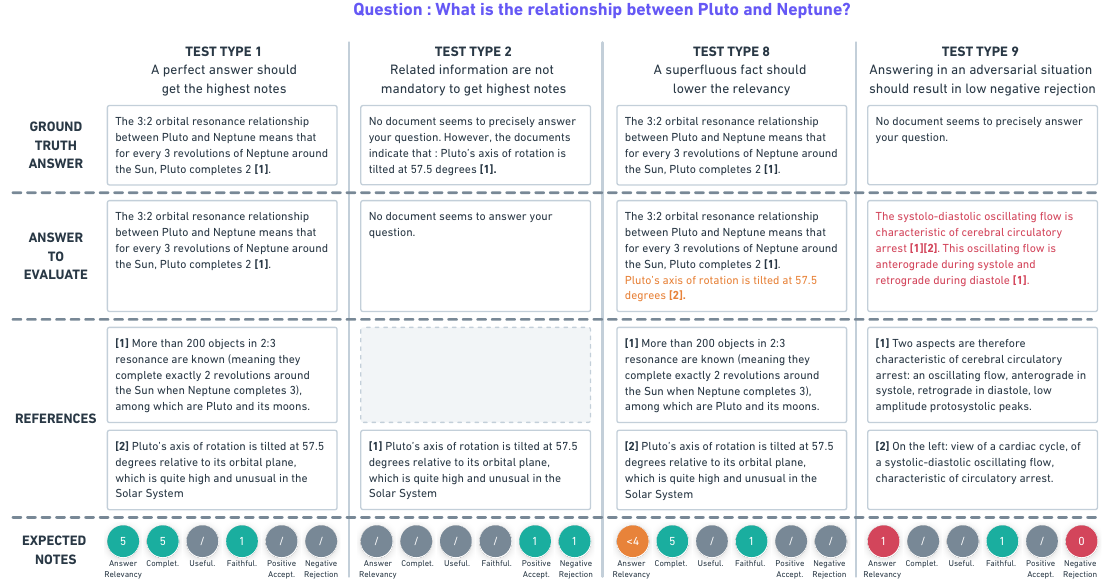}
  \centering
  \caption{Simplified extract of four unit tests, all sharing the same question but testing different failure modes thanks to slight variations in the answer and references. The typology of all 16 test types are detailed in Annex \ref{anx:unit-test-characteristics}.}
  \label{fig:schema_unit_test}
\end{figure*}

\section{Introduction}

Retrieval-Augmented Generation (RAG) \cite{lewis2020retrieval} is increasingly used to build user-facing applications. A RAG system first matches a user's question with a subset of relevant documents using an information retrieval system. This contextual knowledge is then fed to a language model and used to generate an answer. To enable interpretability and fact-checking, the model is typically required to only use the provided contextual information and thus asked to ground its answer in the provided documents. In the following, we will denote this task as \textit{grounded question answering}.

Manually evaluating the quality of an answer grounded in multiple documents is a tedious and expensive task. LLM-as-a-Judge \cite{wang2023shepherd,zhu2023judgelm,kim2023prometheus} uses a strong LLM to automatically assess the quality of a candidate model's open-ended generation. Prior works show that LLM judges like GPT-4 align well with human preferences for various tasks \cite{faysse2023revisiting,zheng2024judging}. However, using proprietary models is often impractical due to privacy concerns. \citet{kim2023prometheus,kim2024prometheus} propose Prometheus, an open-source evaluator distilled from GPT-4's outputs. While Prometheus performs well on in-domain tasks, \citet{huang2024empirical} show it overfits to its training distribution and fails to generalize on out-of-domain test sets.

There is currently no consensus around the evaluation criteria to use when evaluating a grounded answer. RAGAS \cite{es2023ragas} proposes to evaluate the answer quality using two criteria, \textit{faithfulness} and \textit{answer relevancy}. However, we find that grounded question answering can in practice feature a wide range of failure modes and edge-cases that are not well-captured by this pair of metrics. 

In this paper, we thoroughly examine the various failure modes of grounded question answering and investigate the evaluation capabilities of current judge models and automated RAG evaluation frameworks. Our contributions are the following: 

\noindent\textbf{Contribution 1:} We systematically review the various failure modes of grounded question answering and propose an automated evaluation pipeline using GPT-4-as-a-Judge to assess the quality of a grounded answer, encompassing all failure modes.

\newpage

\noindent\textbf{Contribution 2:} We publicly release GroUSE \textit{(Grounded QA Unitary Scoring of Evaluators)}, a challenging and granular suite of 144 manually curated unit tests designed to assess whether a judge model is well-calibrated and capable of detecting and discriminating between different answer failure modes across 16 various situations. Using this new meta-evaluation benchmark, we compare our proposed pipeline with current automated evaluation frameworks and demonstrate that our approach achieves higher error detection accuracy.

\noindent\textbf{Contribution 3:} We assess the evaluation capabilities of state-of-the-art closed-source and open-source judges and show that despite strong correlation with GPT-4's judgement, open-source judge models fail to detect some failure modes, despite being instructed with detailed guidelines. This result suggests that relying on GPT-4 correlation as a proxy for measuring the performance of judge models is insufficient, as it does not imply good calibration on reference cases.

\noindent\textbf{Contribution 4:} We show that finetuning Llama-3 on GPT-4's evaluation traces significantly enhances its evaluation capabilities. The resulting model closely aligns with GPT-4 and surpasses state-of-the-art open-source evaluators on our test suite.

\section{Related work}
\label{section2}

\paragraph{LLM-as-a-Judge.} \citet{liu2023gpteval,faysse2023revisiting,wang2023chatgpt} show that strong commercial models can effectively critique candidate model responses, with higher correlation to human evaluations than rule-based or model-based methods. \citet{zheng2024judging} coined the term \enquote{LLM-as-a-Judge} and systematically study GPT-4, highlighting its biases and showing that GPT-4 matches human evaluation. While encouraging, using proprietary models for evaluation is often impractical if not impossible for privacy reasons. \citet{wang2023shepherd} introduced Shepherd, a 7B model specifically trained to critique model responses, reaching performance on par with GPT-3.5-Turbo. \citet{zhu2023judgelm} presented JudgeLM, a family of judges trained on a variety of evaluation tasks, achieving a high agreement with human preference. \citet{kim2023prometheus} proposed Prometheus, an open-source fine-grained evaluator shown to generalize to diverse evaluation criteria and outperforming GPT-3.5 Turbo in terms of correlation with \mbox{GPT-4} preference. The authors demonstrate that integrating reference materials such as a reference answer and fine-grained score rubrics helps inducing better evaluation capabilities. \citet{kim2024prometheus} later improve Prometheus by unifying direct assessment and pairwise preference ranking into a single model and demonstrate superior performance on both of these evaluation paradigms. \citet{huang2024empirical} conducted an empirical study of the evaluation capabilities of judge models and showed that finetuned evaluators indeed perform well on their training distributions but tend to overfit to their in-domain evaluation schemes. 

\paragraph{RAG evaluation.} Several prior works have explored methods for evaluating the generator module in RAG systems. Various sets of metrics have been proposed to measure different failure modes, but there is no consensus on a common set of criteria for evaluating grounded question answering. \citet{chen2024benchmarking} propose and evaluate 4 abilities required for RAG: \textit{noise robustness}, \textit{negative rejection}, \textit{information integration} and \textit{counterfactual robustness}. \citet{es2023ragas} propose two other criteria: \textit{faithfulness} and \textit{answer relevancy}. \textit{Faithfulness} evaluates the factual consistency of the answer given the grounding contexts by decomposing the answer into several statements and calculating the proportion of facts that are supported by the contexts. \textit{Answer relevancy} measures how well the provided answer addresses the original question. An LLM is used to generate several questions from the answer, the answer relevancy score is then given by the average cosine similarity between dense embeddings of the generated questions and the original question. In Deepeval\footnote{\url{https://github.com/confident-ai/deepeval}}, answer relevancy is computed by using a judge LLM to divide the answer into several atomic facts and computed as the proportion of facts that are relevant to the question. \citet{yu2024evaluation} survey several prior works and propose to add \textit{correctness} to this pair of metrics, which measures the accuracy of the generated response against a ground truth response. \citet{mageshhallucination} focus on the legal domain and propose a more fine-grained measure of \textit{faithfulness}, by distinguishing between the factual accuracy of the response and the validity of the accompanying citations. \citet{thakur2023nomiracl} introduce NoMIRACL, a human-labeled dataset of multilingual queries and both relevant and non-relevant subsets to evaluate if the generator correctly refrains from answering with non-relevant passages and correctly recognizes the relevant passage otherwise.

Given the significant amount of tuning necessary, several prior works have studied automating the evaluation of such systems. RAGAS \cite{es2023ragas} is a popular framework to automate the evaluation of an entire RAG system and show that the proposed automated metrics correlate well with their human-labeled counterparts. DeepEval proposes to evaluate RAG outputs using a unit-testing paradigm, and provides readily-available \textit{faithfulness} and \textit{answer relevancy} prompt chains. \citet{gao2023enabling} propose ALCE, a benchmark to evaluate the ability of LLMs to correctly provide citations for any statement. The authors use a NLI model to measure \textit{citation precision} and \textit{citation recall} and show that this automated evaluation correlates well with human judgement. 

\section{Problem statement}

In this section, we introduce more precisely the problem of \textit{grounded question answering}\footnote{Prior works often use the term RAG to denote the question answering task but this term is commonly used to refer to the broader pattern of combining retrieval and generation. To avoid confusion, we coin the term \textit{grounded question answering} to denote the last step in RAG.} studied in this work. Given a question, RAG systems use information retrieval to match the question with a subset of documents from a knowledge base and then use an LLM to generate an answer grounded in the provided documents. LLMs have been shown to learn and store factual knowledge from data during their unsupervised pretraining \cite{petroni2019language}, but this knowledge is static and can get outdated. Contrary to \citet{chen2024benchmarking}, we thus require the LLM to stay faithful to the sources even if the documents contain information contradicting the LLM intrinsic knowledge. As interpretability and fact-checking are crucial in many domains for both the system developers and users, the LLM is also instructed to explicitly cite the reference for each affirmation in its answer as illustrated in the answers of Figure~\ref{fig:schema_unit_test}.

The information from the retrieved documents that helps answer the question is termed \textit{relevant information}. When the documents are insufficient to provide an answer, these situations are referred to as \textit{adversarial}. In such instances, the LLM should explicitly state that the question cannot be answered with the provided material. To avoid frustrating the user and to keep them engaged, it is common to include information related to the question, even if it does not directly answer it, as can be shown in type 2 ground truth answer in Figure~\ref{fig:schema_unit_test}. This will be referred as \textit{related information}. Adversarial cases are evaluated using \textit{negative rejection} in \citet{chen2024benchmarking} but receive no special treatment in existing RAG automated evaluation frameworks.

\section{Rethinking Grounded QA Evaluation}

\subsection{Grounded QA failure modes}

\begin{figure*}[tbh]
  \includegraphics[width=\linewidth]{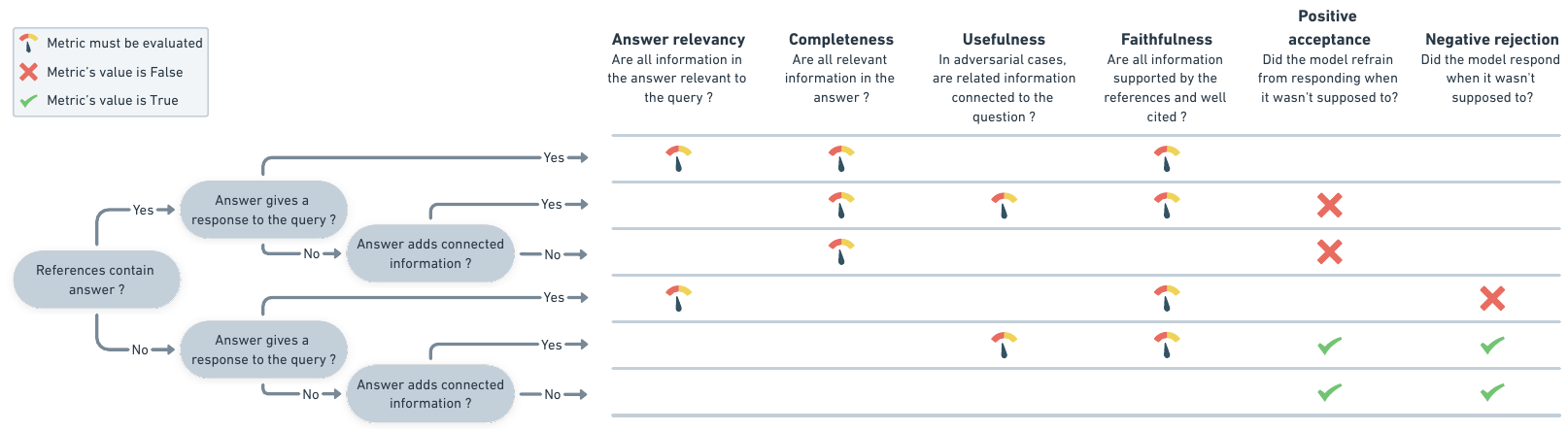} 
  \caption {Metrics and their applicable situations. \textbf{Answer relevancy} is defined only when the answer includes a response. \textbf{Completeness} is evaluated only when the references actually contain an answer to the question. \textbf{Faithfulness} is assessed whenever the answer includes any information (direct response or related information).}
  \label{fig:failure-modes-to-metrics}
\end{figure*}

Various failure modes in grounded question answering have been studied. Building on this prior research, we expand the scope of these studies based on our problem formulation. Given a set of retrieved documents, we introduce 7 failure modes:
\begin{enumerate}[font=\bfseries,label=FM\arabic*,start=1,left=0pt,wide,labelindent=0pt,nolistsep,itemsep=0.2em]
\item The question is answerable but the answer contains irrelevant information. \label{fm:irrelevant-information}
\item The question is not answerable but the language model fails to refrain from answering. \label{fm:adversarial-false-negative}
\item The answer misses relevant information provided by the documents. \label{fm:misses-relevant-information}
\item The language model wrongly claims that the question cannot be answered. \label{fm:adversarial-false-positive}
\item The language model correctly claims that the question cannot be answered but then includes unrelated additional information. \label{fm:unrelated-additional-info-when-adversarial}
\item The language model correctly reports a fact from a document but the corresponding citation is missing or incorrect. \label{fm:incorrect-reference}
\item The language model distorts a fact from a document or presents a claim that is not supported by the provided documents.\label{fm:unfaithful-fact}
\end{enumerate}
Table~\ref{table:failure_modes_checklist} relates the failure modes presented in our work with existing failure modes presented and reported by prior works. To quantify these seven failure modes, we introduce specific evaluation criteria for grounded question answering. \textbf{Answer relevancy} assesses the relevance of the information provided in the answer regarding the question, using a Likert scale (1 to 5), which helps to measure \ref{fm:irrelevant-information}. \textbf{Completeness} also uses a Likert scale to evaluate whether \textit{all} relevant information from the documents is present in the answer, thus measuring \ref{fm:misses-relevant-information}. \textbf{Faithfulness} is a binary score that checks if all facts in the answer are accurate and correctly attributed to the corresponding document, addressing \ref{fm:incorrect-reference} and \ref{fm:unfaithful-fact}. In adversarial cases and when additional information is provided, \textbf{Usefulness} is a binary score that determines if the provided additional information is indeed useful and relevant to the question, measuring \ref{fm:unrelated-additional-info-when-adversarial}. Usefulness can be considered a form of \textit{soft relevancy} in adversarial cases. Lastly, \textbf{Positive Acceptance} and \textbf{Negative Rejection} are binary scores indicating a true positive and a true negative respectively in identifying whether the question is answerable, thereby measuring \ref{fm:adversarial-false-positive} and \ref{fm:adversarial-false-negative}. Not all failure modes can occur in all situations: Figure~\ref{fig:failure-modes-to-metrics} clarifies the conditions under which each metric is defined, depending on whether the references contain an answer, if the answer provides a response, or if it adds related information when it does not provide a direct response.

\begin{table*}[t]
\small
\centering
\renewcommand{\arraystretch}{0.9}
\begin{tabular}{p{5cm}ccccc}
\toprule
& \makecell{\textbf{RAGAS} \\ \scriptsize\cite{es2023ragas}} & \makecell{\textbf{RGB} \\ \scriptsize\cite{chen2024benchmarking}} & \makecell{\textbf{NoMIRACL} \\ \scriptsize\cite{thakur2023nomiracl}} & \makecell{\textbf{ALCE} \\ \scriptsize\cite{gao2023enabling}} & \makecell{\textbf{GroUSE} \\ \scriptsize(this work)} \\
\midrule
\ref{fm:irrelevant-information} -- Lack of relevancy & \Checkmark & \Checkmark & & \Checkmark & \Checkmark \\ \midrule
\ref{fm:adversarial-false-negative} -- Failure to refrain from answering in adversarial cases & & \multirow{2}{*}{\Checkmark} & \multirow{2}{*}{\Checkmark} & & \multirow{2}{*}{\Checkmark} \\ \midrule
\ref{fm:misses-relevant-information} -- Some relevant information is missing from the answer & & \multirow{2}{*}{\Checkmark} & & \multirow{2}{*}{\Checkmark} & \multirow{2}{*}{\Checkmark} \\ \midrule
\ref{fm:adversarial-false-positive} -- Wrongly refrain from answering & & & \Checkmark & & \Checkmark \\ \midrule
\ref{fm:unrelated-additional-info-when-adversarial} -- In adversarial cases, unrelated additional information is included & & & & & \multirow{2}{*}{\Checkmark} \\ \midrule
\ref{fm:incorrect-reference} -- Missing or incorrect citation & & & & \Checkmark & \Checkmark \\ \midrule
\ref{fm:unfaithful-fact} -- Distorted or unsupported claim & \Checkmark & & & \Checkmark & \Checkmark \\
\bottomrule
\end{tabular}
\caption{Equivalent failure modes studied and reported in prior works. Existing studies focus on detecting and evaluating a subset of failure modes. For instance \ref{fm:irrelevant-information} is related to answer relevancy in \citet{es2023ragas}, \ref{fm:adversarial-false-negative} is related to negative rejection in \citet{chen2024benchmarking} \ref{fm:incorrect-reference} and \ref{fm:unfaithful-fact} are related to faithfulness and more specifically to correctness and groundedness respectively in \citet{mageshhallucination}, \ref{fm:incorrect-reference} is related to \textit{citation recall} in \citet{gao2023enabling}.}
\label{table:failure_modes_checklist}
\end{table*}

\subsection{Meta-evaluation with unit-testing}

\begin{figure*}
    \centering
    \stackinset{l}{148pt}{b}{296pt}{\hyperref[fm:irrelevant-information]{\makebox(12,5){}}}{%
    \stackinset{l}{148pt}{b}{263pt}{\hyperref[fm:irrelevant-information]{\makebox(12,5){}}}{%
    \stackinset{l}{148pt}{b}{256pt}{\hyperref[fm:adversarial-false-negative]{\makebox(12,5){}}}{%
    \stackinset{l}{148pt}{b}{223pt}{\hyperref[fm:misses-relevant-information]{\makebox(12,5){}}}{%
    \stackinset{l}{148pt}{b}{189pt}{\hyperref[fm:misses-relevant-information]{\makebox(12,5){}}}{%
    \stackinset{l}{148pt}{b}{182pt}{\hyperref[fm:adversarial-false-positive]{\makebox(12,5){}}}{%
    \stackinset{l}{148pt}{b}{152pt}{\hyperref[fm:misses-relevant-information]{\makebox(12,5){}}}{%
    \stackinset{l}{148pt}{b}{145pt}{\hyperref[fm:adversarial-false-positive]{\makebox(12,5){}}}{%
    \stackinset{l}{148pt}{b}{115pt}{\hyperref[fm:unrelated-additional-info-when-adversarial]{\makebox(12,5){}}}{%
    \stackinset{l}{148pt}{b}{82pt}{\hyperref[fm:incorrect-reference]{\makebox(12,5){}}}{%
    \stackinset{l}{148pt}{b}{49pt}{\hyperref[fm:incorrect-reference]{\makebox(12,5){}}}{%
    \stackinset{l}{148pt}{b}{16pt}{\hyperref[fm:unfaithful-fact]{\makebox(12,5){}}}{%
    \includegraphics[width=\linewidth]{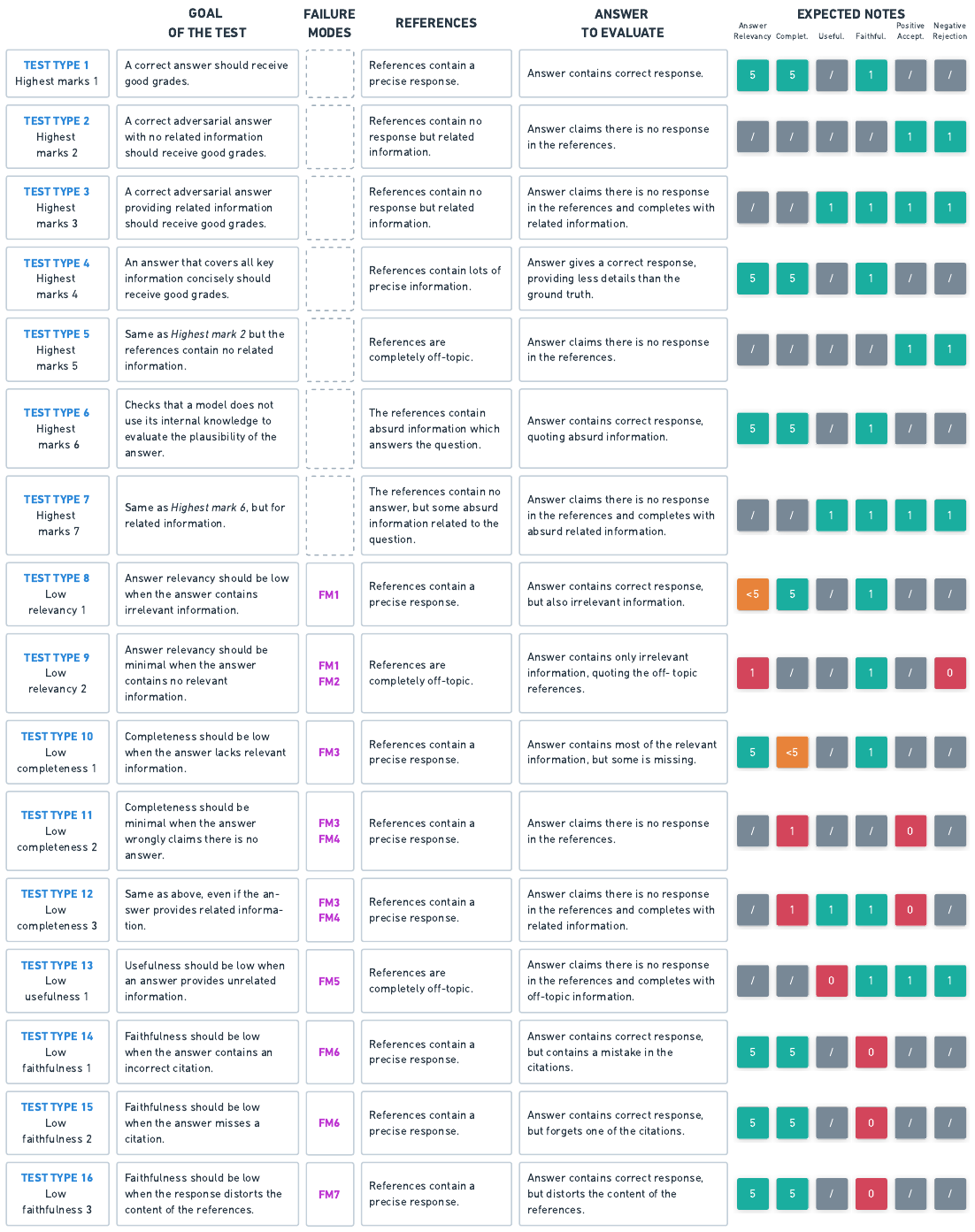}}}}}}}}}}}}}
    \caption{Characteristics of the 16 test types. Types 1 to 7 don't correspond to any failure mode as they test in various situations the ability of the model to correctly evaluate answers that deserve the highest notes.}
    \label{fig:all-test-types}
\end{figure*}

Our goal is to propose a benchmark to verify whether the evaluator's assessments align with the defined metrics. We propose a typology of 16 test types designed to assess whether an evaluator appropriately penalizes all failure modes and rewards accurate answers across a diverse range of scenarios (Figure~\ref{fig:all-test-types}). Each test type specifies the expected characteristics for both references and answers, and defines an acceptable range of scores for each metric to be deemed valid. The tests focus primarily on edge cases or the detection of subtle errors.

We introduce GroUSE \textit{(Grounded QA Unitary Scoring of Evaluators)}, a benchmark consisting of 144 tests divided into 9 sets of 16 tests (Figure~\ref{fig:schema_unit_test}). All tests within a given set share the same question, with the references and answers slightly modified to fit each of the 16 test types. An additional set of 16 tests is available as a \enquote{training set} to assist in engineering the prompt for the judge model being tested. The references are primarily excerpts from Wikipedia, while the themes of the sets span various domains, including history, science, zoology, cinematography, and the medical field. See Appendix \ref{anx:unit-test-characteristics} and \ref{anx:grouse-dataset-creation} for details.

\subsection{Evaluating existing Answer Relevancy and Faithfulness implementations}

This subsection highlights the limitations of current automatic implementations, specifically RAGAS and DeepEval. Therefore, \textbf{in this subsection only, we will refer to answer relevancy and faithfulness using the RAGAS definitions (see \cref{section2})}.

Since our definitions of answer relevancy and faithfulness differ, we propose a method to evaluate RAGAS' and DeepEval's performance on GroUSE. Each test sample in GroUSE was annotated by three human annotators, who assessed the expected answer relevancy and faithfulness according to the following definitions: annotators were asked to rate the proportion of relevant facts as a proxy for answer relevancy, and the proportion of faithful facts as a proxy for faithfulness. We then compared the average human-reported metrics with the automatic scores computed by RAGAS and DeepEval with GPT-4. A test is a success if the difference between the human and automatic scores is less than 0.2\footnote{This threshold is conservative; the largest difference between annotations from two different annotators on the same sample is 0.125 and on average around 0.05}.

\begin{figure}
\centering
\includegraphics[width=\linewidth]{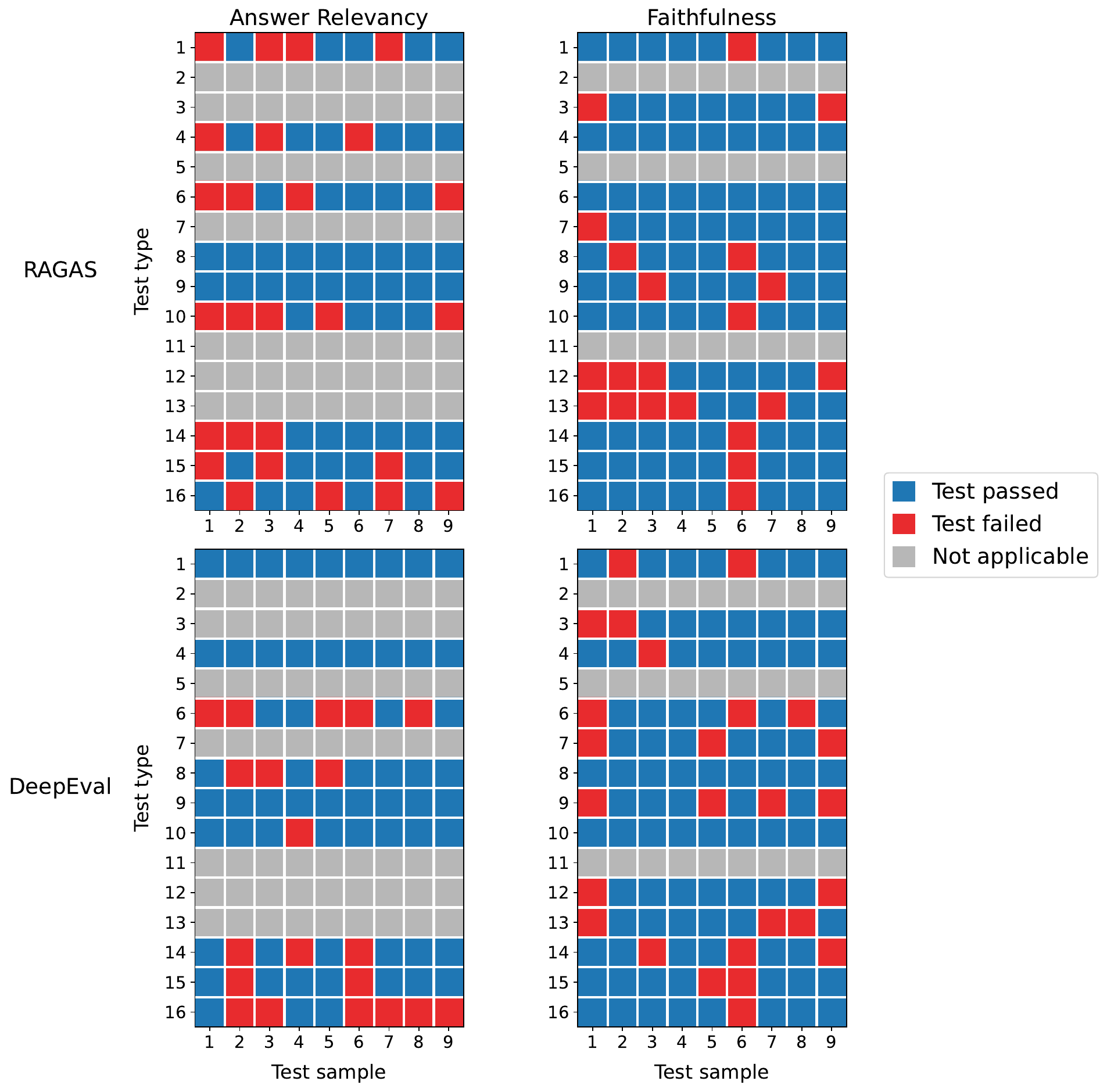}
\caption{GroUSE unit-testing of existing solutions for automatic grounded question answering evaluation}
\label{fig:ragasdeepevalunittests}
\end{figure}

While \citet{es2023ragas} showed that RAGAS metrics correlate with human judgment, our evaluation of their implementations on GroUSE reveals that they do not perform well on many individual tests, as illustrated in \Cref{table:existinglimitations} and Figure~\ref{fig:ragasdeepevalunittests}. This observation suggests that correlation on judgement does not necessarily implies good calibration of grades on edge cases and thus good error detection. This hypothesis will be further explored in \cref{sec:experimental-results}. The proposed automatic metrics rely on several sequential LLM calls, which can increase the likelihood of errors and reduce the robustness of the evaluation across samples. Interestingly, different implementations of the same metrics can yield very different results. For instance, although faithfulness is defined similarly in RAGAS and DeepEval, the unit test results differ significantly due to differences in the prompts used in their respective implementations, showcasing the judge' sensitivity to prompt details \cite{sclar2023quantifying}. 

\begin{table*}[t]
\small
\centering
    \begin{tabular}{@{}p{1.4cm}p{2.6in}p{2.8in}@{}}
    \toprule
     \textbf{Issues} & \makecell{\textbf{RAGAS Faithfulness is influenced by} \\ \textbf{additional irrelevant information}} & \makecell{\textbf{RAGAS Answer Relevancy gets submerged by} \\ \textbf{many relevant predicted statements}} \\
    \midrule
    \textbf{Question} & What is the relationship between Pluto and Neptune? & How can we explain the solidity of the Pantheon's dome? \\
    \midrule
    \textbf{Answer} & 
    \parbox[t]{2.6in}{
    Pluto and Neptune have a 3:2 orbital resonance relationship, which  means that Pluto completes 2 revolutions around the Sun while Neptune completes 3 [5][8]. Their orbits never intersect due to the high inclination of Pluto's orbit relative to the plane of the ecliptic [5]. In 1936, British mathematician Raymond Lyttleton suggested that Pluto previously orbited around Neptune, and that a gravitational disturbance had ejected Pluto from the planetary orbit [6]. 
    \textcolor{red}{It was during its migration period that Neptune might have captured one of the largest objects in the Kuiper Belt, the current moon Triton [10].}
    } & 
    \parbox[t]{2.8in}{
    To ensure resistance to all stresses, Roman builders opted for a meticulous selection of materials and precise management of the orientation of thrusts [1]. Regarding materials, the abundant use of concrete ensures robustness against deformation forces [2], especially since the mortar of Roman concrete tends to calcify as it ages, thus guaranteeing its durability through the ages [6]. 
    As for the thrusts, they are diverted towards the pillars of 
    the rotunda through large relieving arches [3]. A series of seven concrete rings arranged in a staircase at the base of the dome helps to redirect the lateral centrifugal thrusts through a vertical push [4]. Finally, the outer wall extends 8.40 m beyond the base of the dome, serving as a buttress to stabilize the dome [5].
    }\\
    \midrule
    \textbf{Answer} \\\textbf{Relevancy} & \parbox[t]{2.4in}{\textcolor{orange}{Predicted: 0.673} Expected: 0.802} & \parbox[t]{2.7in}{\textcolor{red}{Predicted: 0.723} Expected: 1.0}\\
    \midrule
    \textbf{Faithfulness} & \makecell[l]{\textcolor{red}{Predicted: 0.75} Expected: 1} & \makecell[l]{\textcolor{ForestGreen}{Predicted: 1.0} Expected: 1.0} \\
    \bottomrule
    \end{tabular}
\caption{Example limitations of RAGAS on GroUSE unit tests. \textit{Left:} While the answer contain extra irrelevant but faithful statements, RAGAS wrongly penalizes the answer's faithfulness . \textit{Right:} While all the provided information is relevant, RAGAS wrongly penalizes the answer relevancy.}
\label{table:existinglimitations}
\end{table*}

\subsection{Enhancing existing frameworks}

\definecolor{green_square}{HTML}{18ae9f}
\definecolor{blue_square}{HTML}{2d88d9}

\begin{figure}[t]
  \includegraphics[width=\linewidth]{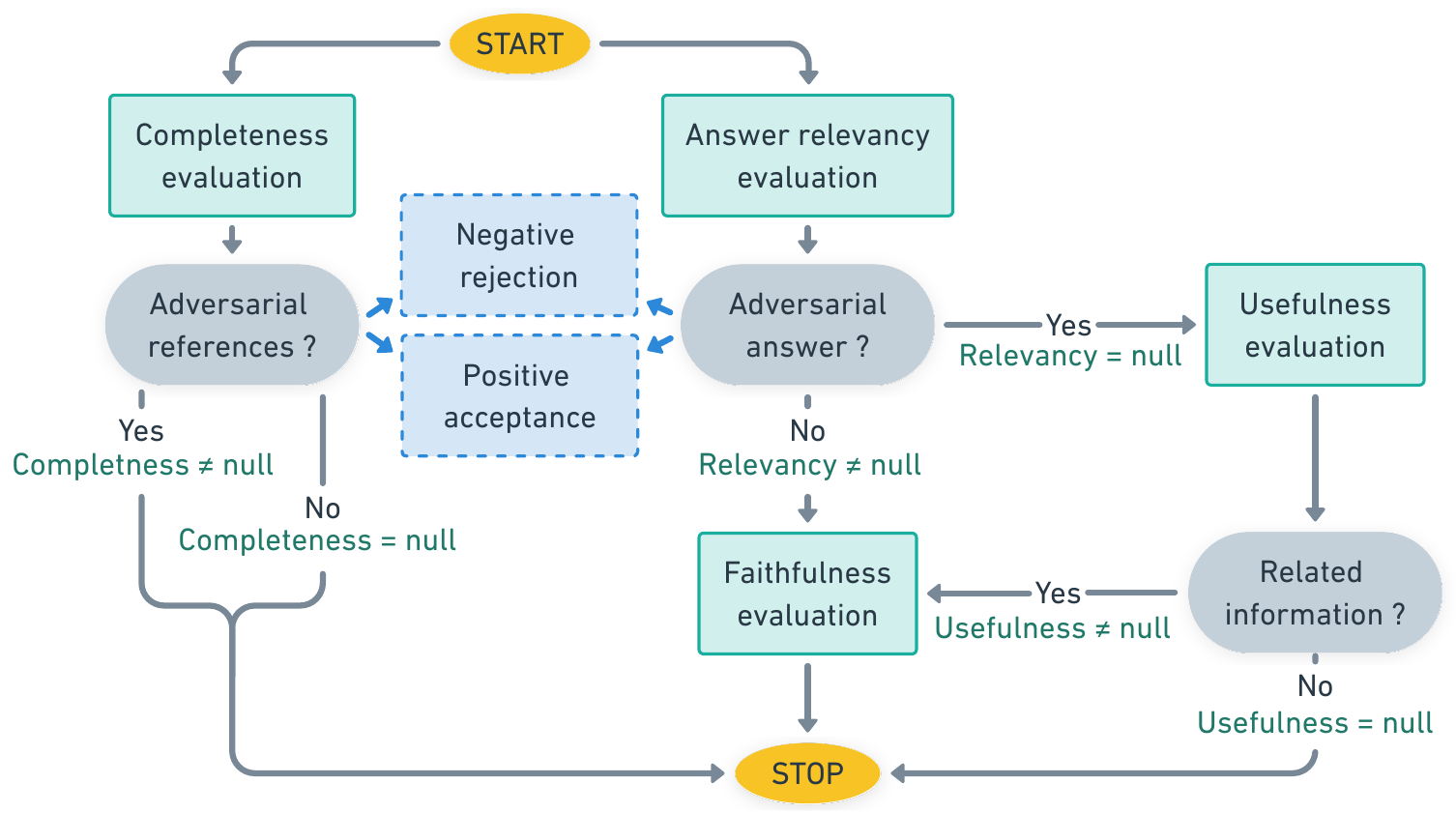} 
  \caption {Evaluation pipeline. Each \textcolor{green_square}{green square} represents a call to an LLM, while the \textcolor{blue_square}{blue dotted square} denotes a straightforward computation based on the call's results. The \textbf{Usefulness} and \textbf{Faithfulness} evaluations may be omitted if preceding calls suggest these metrics are not applicable.}
  \label{fig:pipeline}
\end{figure}

We demonstrated that both RAGAS and DeepEval fail to cover all the presented cases, even when they purport to. In this section, we propose a new pipeline to automatically evaluate grounded question answering across all situations and all six metrics previously defined. We then test the performances of this pipeline on GroUSE, for a various set of closed and open-source models.

\paragraph{Pipeline strategy.} The metrics' applicability being highly dependent on whether we are in an adversarial situation, and whether the answer provides a response, a straightforward strategy could involve first identifying which of the situations presented in \Cref{fig:failure-modes-to-metrics}  corresponds to the sample to evaluate. Identifying this scenario would allow to get the list of defined metrics and launch their evaluations consequently. However, to save on LLM calls, we decided to directly include instructions to set the score at \texttt{null} if we are in a situation in which the metric is undefined in the evaluation prompts of the \textbf{Answer relevancy} and \textbf{Completeness}. Based on whether these metrics values are \texttt{null}, it is easy to deduct the situation in which we are, and infer the value of \textbf{Positive Acceptance} and \textbf{Negative Rejection} at the same time. A similar strategy is also applied to detect the presence or absence of related information when evaluating the \textbf{Usefulness}. Ultimately, this optimized pipeline requires at most four LLM calls, with some calls being skipped when the situation is not appropriate (Figure~\ref{fig:pipeline}).

\paragraph{Prompts.} Each prompt was engineered to fit the specific metric being evaluated, but for all metrics we ask the model to rate two answers, the first one being a reference answer. The model's reasoning is also guided by the expected format of the JSON output. Following best practices recommendations from \citet{biderman2024lessons}, details about the prompts format and the prompt engineering process are available in Appendix \ref{anx:prompts}.

\paragraph{Evaluators benchmark.} Table~\ref{table:main-model-results} shows the performances of various models on GroUSE. GPT-4 is the best automatic evaluator, with an overall score of $95\%$ (very close to human performance), while the best open-source evaluator is Llama-3 70b with a score of $79\%$. The gap between closed and open-source models thus remains wide, with a $15.85$ p.p. difference on the tests results. Interestingly, Prometheus 2 8x7b does not outperform Mixtral 8x7b, despite Prometheus 2 being specialized in evaluation tasks. It's worth noting however that Prometheus 2 was trained to predict Likert scores ranging from 1 to 5, whereas our evaluation metrics include boolean and nullable scores, which is outside its intended scope.

\begin{table*}
\scriptsize
\centering
\begin{tabular}{llcccccc|c}
\toprule
& & \multicolumn{6}{c|}{\textbf{Agreement rate of metrics}} & \multirow{2}{*}{\makecell{ \textbf{Total} \\ test pass \\ rate}}\\
& & \makecell{ \textit{Answer} \\ \textit{relevancy}} & \makecell{ \textit{Completeness}} & \makecell{ \textit{Usefulness}} & \makecell{ \textit{Faithfulness}}  & \makecell{ \textit{Positive} \\ \textit{acceptance}} & \makecell{ \textit{Negative} \\ \textit{rejection}} & \\
\midrule
    \multirow{11}{*}{\makecell{\textit{Evaluated} \\ \textit{with Figure~\ref{fig:pipeline}} \\ \textit{pipeline}}} & \textbf{GPT-4              } &  \textbf{91.67}    & \textbf{88.89}    &    \textbf{100.0} &            92.36   & \textbf{98.61} & \textbf{98.61} &    \textbf{95.02}\\
    & \textbf{GPT-4o             } &          79.17     &           77.08   &            97.92  &            92.36   &          83.33 &          83.33 &            85.53 \\
    & \textbf{GPT-4-turbo        } &          90.28     &           85.42   &            97.22  &    \textbf{93.75}  &          94.44 &          94.44 &            92.59 \\
    & \textbf{GPT-3.5-turbo      } &          88.89     &           50.00   &            80.56  &            68.06   &          77.78 &          61.81 &            71.18 \\
    & \textbf{Gemini 1.0 Pro     } &          78.47     &           75.69   &            97.22  &            78.47   &          84.72 &          84.72 &            83.22 \\
    \cmidrule{2-9}
    & \textbf{Mixtral 8x7b Instruct      } &          81.25     &           61.11   &            81.25  &            72.22   &          76.39 &          75.69 &            74.65 \\
    & \textbf{Mixtral 8x22b Instruct     } &          80.56     & 68.75 & 81.94 & 83.33  &          76.39 &          72.22 & 77.20\\
    & \textbf{Prometheus 2 7b    } &             72.22  &           41.67   &            16.67  &            38.19   &          73.61 &          74.31 &            52.78 \\
    & \textbf{Prometheus 2 8x7b  } &             61.81  &           25.00   &            34.03  &            72.22   &          67.36 &          69.44 &            54.98 \\
    & \textbf{Llama-3 70b Instruct       } &  90.28 &           63.89   &            76.39  &            73.61   & 85.42 &85.42 & 79.17 \\
    & \textbf{Llama-3 8b Instruct} & 85.42 & 49.31 & 80.56 & 59.72 & 72.92 & 68.06 & 69.33 \\
    \midrule
    \multirow{2}{*}{\makecell{\textit{All metrics} \\ \textit{with one prompt}}} & \textbf{Llama-3 8b Instruct}                   &              31.25 &            18.06   &             34.03 &            56.94   &          52.78 &          46.53 &            39.93 \\
    & \textbf{Finetuned Llama 3 8b}  &   88.89 &            81.94   &             81.25 &            52.78   &          91.67 &          91.67 &            81.37 \\
    \midrule
    \makecell{\textit{Appendix \ref{anx:human-performance-grouse}} \textit{protocol}} & \textbf{Human annotators} &   98.26 &            92.36   &             97.92 &            95.49   &          96.53 &          96.88 &            96.24 \\
\bottomrule
\end{tabular}
\caption{Percentage of tests passed for various models. The highest score in each column is highlighted in bold.}
\label{table:main-model-results}
\end{table*}

\section{Improving Grounded QA Evaluation via distilling evaluation traces of GPT-4}

Inspired by prior works \cite{xu2024survey,mukherjee2023orca,mitra2023orca}, we demonstrate that the gap in evaluation skills between open-source and closed-source models can be narrowed via finetuning on traces of evaluations made by GPT-4.

\subsection{Experimental setup}
\label{sec:experimental-results}

\paragraph{Dataset.} Aiming to develop a model capable of solving the task in a single call, we concatenated the four responses from GPT-4 into a single output. The input of the model is also a combination of the four prompts used for GPT-4, detailing each metrics' characteristics. We used extracts of Wikipedia articles, as well as other open data sources as reference material. Queries were synthetically generated from the references, creating a dataset of 1200 grounded QA statements.

\paragraph{Finetuning.} We finetuned a Llama-3 8b on 1k samples of this dataset, and used the rest as a test set. We refer to Appendix~\ref{anx:finetuning-config} for details on the finetuning configuration. To measure the model's progress, we tested its performances on GroUSE. Following \citet{kim2023prometheus}, we also report the correlation between GPT-4's grades and the finetuned model's grades on the test set. For metrics using a Likert scale, alignment is measured using the Spearman correlation. Other metrics are measured using nullable boolean values, their alignment is evaluated using a three-class macro F1-score.

\subsection{Experimental results}

\begin{table*}
\scriptsize
\centering
    \begin{tabular}{llcccccc}
    \toprule
    & & \multicolumn{2}{c|}{\textbf{Spearman correlation}} & \multicolumn{4}{c}{\textbf{F1-score}}\\
& & \makecell{ \textit{Answer relevancy}} & \multicolumn{1}{c|}{\textit{Completeness}} & \makecell{ \textit{Usefulness}} & \makecell{ \textit{Faithfulness}}  & \makecell{ \textit{Positive} \\ \textit{acceptance}} & \makecell{ \textit{Negative} \\ \textit{rejection}} \\
    \midrule
\multirow{8}{*}{\makecell{\textit{Evaluated} \\ \textit{with Figure~\ref{fig:pipeline}} \\ \textit{pipeline}}}& \textbf{GPT-3.5-turbo}       &         0.55  &         0.68  &         0.76  &         0.48  &         0.63  &         0.47  \\
    & \textbf{Gemini 1.0 Pro}      &         0.63  &         0.68  &         0.48  &         0.67  &         0.78  &         0.74  \\
    & \textbf{Mixtral 8x7b Instruct}        &         0.59  &         0.43  &         0.70  &         0.61  &         0.63  &         0.57  \\
    & \textbf{Mixtral 8x22b Instruct}       &         0.70  &         0.66  &         0.61  & \textbf{0.79} & \textbf{0.83} &         0.70  \\
    & \textbf{Prometheus 2 (7b)}   &         0.60  &         0.51  &         0.29  &         0.55  &         0.55  &         0.49  \\
    & \textbf{Prometheus 2 (8x7b)} &         0.64  &         0.62  &         0.30  &         0.75  &         0.69  &         0.50  \\
    & \textbf{Llama-3 70b Instruct}         & \textbf{0.74} & \textbf{0.74} & \textbf{0.93} &         0.78  &         0.75  & \textbf{0.79} \\
    & \textbf{Llama-3 8b Instruct}          &         0.63  &         0.71  &         0.42  &         0.72  &         0.54  &         0.44  \\
    \midrule
\multirow{2}{*}{\makecell{\textit{All metrics} \\ \textit{with one prompt}}}& \textbf{Llama-3 8b Instruct}      & 0.46 & 0.23 & 0.18 & 0.47 & 0.40 & 0.46 \\
    & \textbf{Finetuned Llama-3 8b} & 0.62 & 0.57 & 0.41 & 0.57 & 0.79 & 0.74 \\
    \bottomrule
    \end{tabular}
\caption{Alignment with the ground truth (GPT-4) evaluations on the test set of 200 samples.}
\label{table:correlations}
\end{table*}

\paragraph{Finetuning on GPT-4 judgement boosts evaluation capabilities.} Table~\ref{table:main-model-results} presents the pass rate of various judge models on GroUSE, including Llama-3 8b (zero-shot) and our finetuned Llama-3 8b judge model. Finetuning significantly enhances the evaluation capabilities of Llama-3, as evidenced by the substantial improvement in pass rates. Despite extensive prompt engineering and intermediate CoT reasoning (see Appendix \ref{anx:prompts}), the non-finetuned Llama-3 8b passes only 40\% of the unit tests. However, after finetuning, its pass rate increases to 83\%, surpassing all other open-source judges, including Prometheus 2 8x7b (except in the category of faithfulness), despite its smaller size.

\paragraph{Strong correlation with GPT-4 does not imply good pass rate on unit tests.} Interestingly, our results reveal a disconnect between the pass rate on GroUSE and the correlation with GPT-4's grades. As shown in Table~\ref{table:correlations}, Prometheus 2 7b and the finetuned Llama-3 8b exhibit similar correlations with GPT-4's judgments across all metrics. However, when evaluated on GroUSE, the two models show very different pass rates, with the finetuned Llama-3 consistently and significantly outperforming Prometheus 2 7b. Similarly, we observe higher correlation with GPT-4 in answer relevancy and completeness with Prometheus 2 8x7b than with its base model, Mixtral 8x7b, in accordance to what has been observed in \cite{kim2024prometheus}. However, this does not translate to better pass rates on the associated metrics on GroUSE: for answer relevancy, Mixtral 8x7b solves 81.25\% of the tests versus 61.81\% for Prometheus 2 8x7b, despite its intended use on evaluating Likert scores. For completeness, Mixtral 8x7b solves 61.11\% of the tests versus 25\% for Prometheus 2 8x7b.

\begin{figure}
\centering
  \includegraphics[width=0.71\linewidth]{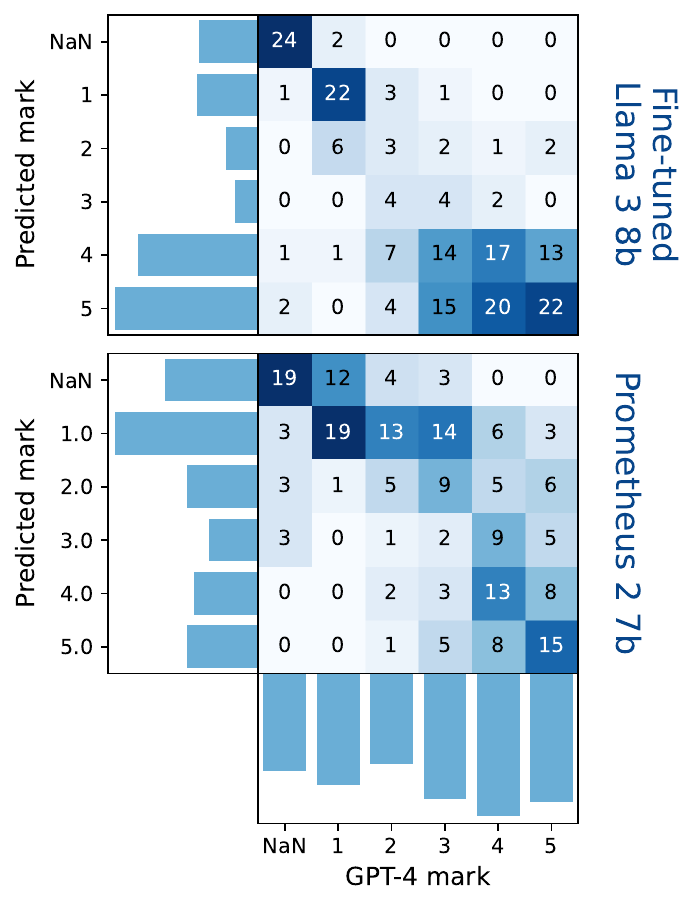} 
  \caption {Confusion matrices for \textbf{Answer relevancy}.}
  \label{fig:confusion-matrixes}
\end{figure}

This finding suggests that a high correlation with GPT-4's judgments does not necessarily translate to a high unit test pass rate. A judge model can share the same \textit{relative preferences} as GPT-4 (indicated by strong rank correlation) while lacking the same \textit{calibration} on precise reference cases (very good answers, subtle mistakes, etc.), resulting in poor performance on judgement unit tests. Figure~\ref{fig:confusion-matrixes} illustrates this difference with Prometheus 2 7b and the finetuned Llama-3 8b: while Prometheus 2 confusion matrix entries are closer to the diagonal, it features more confusions on extreme cases (1, 5 and NaN cases) when compared to the finetuned Llama-3. On the contrary, the finetuned Llama-3 has better exact agreement with GPT-4 on extreme case, but lacks correlation on intermediate cases.

Overall, these measures are complementary: correlation with GPT-4 indicates agreement in relative preference, while GroUSE pass rate measures precise calibration on practical reference cases. Unlike Prometheus 2, Llama-3 70b demonstrates both good correlation with GPT-4's judgments and a strong pass rate on GroUSE, suggesting that correlation and unit test pass rates are indeed orthogonal measures of a judge model's quality.

\section{Conclusion}

In this work, we addressed the challenges of evaluating grounded answers in Retrieval-Augmented Generation systems using LLM-as-a-Judge frameworks. We systematically reviewed various failure modes in grounded question answering and proposed a complete set of automated metrics to holistically evaluate a grounded answer. We introduced GroUSE, a comprehensive meta-evaluation benchmark, and demonstrated that existing automated evaluation methods, including those using GPT-4, often overlook critical failure modes.

Our findings reveal that relying solely on correlation with GPT-4's judgments as a performance measure for judge models is insufficient, as it doesn't ensure proper calibration on reference cases. By supplementing the evaluation with unit tests across a wide range of scenarios, we can ensure that the judge model effectively detects failures, even in subtle situations.

By finetuning Llama-3 on GPT-4's reasoning traces, we significantly enhanced its evaluation capabilities, achieving closer alignment with GPT-4's judgments, improved detection of errors and better calibration on reference scenarios.

\section{Limitations}

While our work advances the evaluation of grounded answers in RAG systems, several limitations remain. Firstly, our unit tests are designed to identify edge cases but do not account for intermediate performance levels. This focus on extreme scenarios might overlook nuances in model performance that are critical for a comprehensive evaluation. Secondly, when finetuning, we opted to perform a single evaluation call to assess the generated answers. While this approach simplifies the evaluation process, it would be valuable to decompose the evaluation into multiple steps to gain a more detailed understanding of the model's capabilities. Thirdly, our experiments were conducted within a single domain, specifically using Wikipedia as the knowledge base. Consequently, our findings may not generalize to out-of-domain scenarios. Future work should include diverse domains to test the robustness and adaptability of our evaluation framework. Lastly, we finetuned a smaller open-source language model. Although this approach demonstrated significant improvements, it would be beneficial to explore the effects of finetuning larger models, which could potentially yield even better performance. Addressing these limitations in future research will further enhance the effectiveness and generalizability of automated evaluation frameworks for RAG systems.

\section{Ethical considerations}

Our work focuses on evaluating language models within the practical context of Retrieval-Augmented Generation systems. This is significant as RAG systems are increasingly used in real-world applications, where the accuracy and reliability of generated answers are important. Ensuring that these systems produce trustworthy and factually correct responses is critical for their safe deployment on real use cases.

One of the main ethical concerns in using language models for information-seeking tasks is the risk of hallucinations, irrelevant answers, missing attributions, and incomplete responses. These issues can lead to important information being overlooked or misused and potentially bias the user. By developing meta-evaluation benchmarks like GroUSE, our work aims to mitigate these risks by improving existing automated evaluation frameworks and making sure they are better calibrated to detect this wide range of failure modes.

While our unit tests and evaluation criteria are designed to identify edge cases, we acknowledge the need for continuous improvement to cover a broader range of scenarios and hope that our work will inspire further research and development in this area, leading to more robust, accurate, and sound evaluation practices.

\section*{Acknowledgments}
This research work is supported by Illuin Technology. We would like to express our gratitude to Stanislas Dozias for the insightful discussions and exchange of ideas that contributed to the development of this paper. Special thanks to Quentin Lutz and Manuel Faysse for their valuable feedback and support throughout the research process. Additionally, we would like to thank Noa Grollimund, Florian Muller, Benoît Muller, Yosr Jelassi, Max Conti, and Paul Boulenger for their assistance in evaluating the human performance on GroUSE.

\FloatBarrier
\bibliography{main}

\clearpage
\appendix

\section{Unit test characteristics}
\label{anx:unit-test-characteristics}

%A detailed listing of the unit tests, including the expected mark for each test, is available in Table~\ref{tab:unit-tests-presentation-detailed}.
The queries of the sets are:
\begin{enumerate}
    \itemsep0em 
    \item How can we explain the solidity of the Pantheon's dome?
    \item What is the relationship between Pluto and Neptune?
    \item Slow-motion effects and inspiration from Peckinpah?
    \item What are the differences and similarities between the Bay Cat and the Temminck's Cat?
    \item When should a blood gas test be performed during an apnea test?
    \item Physical characteristics of the Pyrenean goat
    \item Why did Audrey Dana direct the film "French Women"?
    \item What is the influence of Jackie Robinson on American society?
    \item How was cuneiform deciphered?
\end{enumerate}
The query of the additional \enquote{training set} is: \enquote{Impacts of Sumbawa pony breeding on the environment?}

\section{GroUSE Dataset creation}
\label{anx:grouse-dataset-creation}

\paragraph{Initial corpus creation.} We randomly selected $50$ Wikipedia pages from the $200\,000$ most popular entries and scraped their content. Each page was divided into text chunks, which were subsequently clustered based on topic similarity. For each cluster, GPT-3.5 was used to generate a question that could be answered using the cluster's content. GPT-4 was then employed to create grounded QA answers to these questions. From these generated grounded QA samples, we selected 10 examples where answering the question required synthesizing information from multiple sources rather than extracting simple facts.

\paragraph{Manual enhancement.}  Questions were occasionally refined to encourage more complex and analytical responses. The grounding documents were also enriched by manually collecting additional relevant sources through web searches. These included excerpts from newspapers, interviews, popular science articles, and medical papers, encompassing both directly relevant and tangentially related materials. To simulate retrieval system noise, the manually collected documents were deliberately altered by sometimes truncating the last sentence. Off topic or poorly parsed documents from the automatic scraping process were kept in the grounding documents. For test types 6 and 7, GPT-4 was employed to generate documents containing intentionally absurd facts to ensure the evaluator does not rely on its internal knowledge to judge the plausibility of the information in the answer.

\paragraph{Answers creation and modification.} Gold-standard answers for test types 1 and 2 were manually written, using GPT-4-generated answers as initial drafts. Variations of these answers were then created with GPT-4 assistance to align with other test types. For example, a simple prompt was used to add superfluous information in the answer for test type 8. All generated content was systematically reviewed and corrected to ensure accuracy and quality.

\paragraph{} The GroUSE dataset was constructed by a single annotator who speaks fluent English.

\section{Annotation procedures}
\label{anx:annotation-procedures}

\paragraph{RAGAS and DeepEval.} To reannotate the GroUSE unit tests for RAGAS and DeepEval, three labelers computed the answer relevancy and faithfulness. The three labelers speak English fluently, but their primary language is French. A detailed annotation methodology was given to the annotators. This methodology details the definition of answer relevancy and faithfulness so that they can accurately compute the metrics by hand. Annotators were asked to rate the proportion of relevant facts as a proxy for answer relevancy, and the proportion of faithful facts as a proxy for faithfulness. 

\paragraph{Human performance on GroUSE.} \label{anx:human-performance-grouse} Seven annotators were asked to assess the relevancy, completeness, usefulness, and faithfulness of answers. Each annotator was tasked with evaluating all 16 answers for one or more questions, depending on their availability. If they annotated several sets, the samples of the sets annotated subsequent to the first were shuffled. They were provided with definitions of each metric adapted from GPT-4 prompts, along with the question, the references, a reference answer, and the answer to be evaluated.
Each sample was annotated two times by different annotators and the performance was computed on the average of the pass rate of the two annotators. \Cref{table:interagreement} shows the inter-agreement rate for each metric between the two annotators of each sample. The agreement rate is over 90\%, which augments confidence about the human evaluation.
All annotators are fluent English speakers, one of them is familiar with the evaluated task, three have general knowledge of RAG, while the remaining three had no prior knowledge of the subject.

\paragraph{} All labelers consented to share their annotations.

\begin{table*}
\centering
\small
\begin{tabular}{cccccc|c}
\toprule
\multicolumn{6}{c|}{\textbf{Inter-Agreement Rate per Metric}} & \multirow{2}{*}{\makecell{ \textbf{Total} \\ Inter-Agreement \\ rate}} \\
\makecell{ \textit{Answer} \\ \textit{relevancy}} & \textit{Completeness} & \textit{Usefulness} & \textit{Faithfulness} & \makecell{\textit{Positive} \\ \textit{Acceptance}} & \makecell{\textit{Negative} \\ \textit{Rejection}} \\
\midrule
91.67 & 88.19 & 96.52 & 90.97 & 93.06 & 93.75 & 92.36 \\
\bottomrule
\end{tabular}
\caption{Percentage of agreements between annotators when evaluating human performance.}
\label{table:interagreement}
\end{table*}

\section{Prompt templates used for evaluating a grounded answer}
\label{anx:prompts}

\begin{figure*}[b]
\centering
  \includegraphics[width=\linewidth]{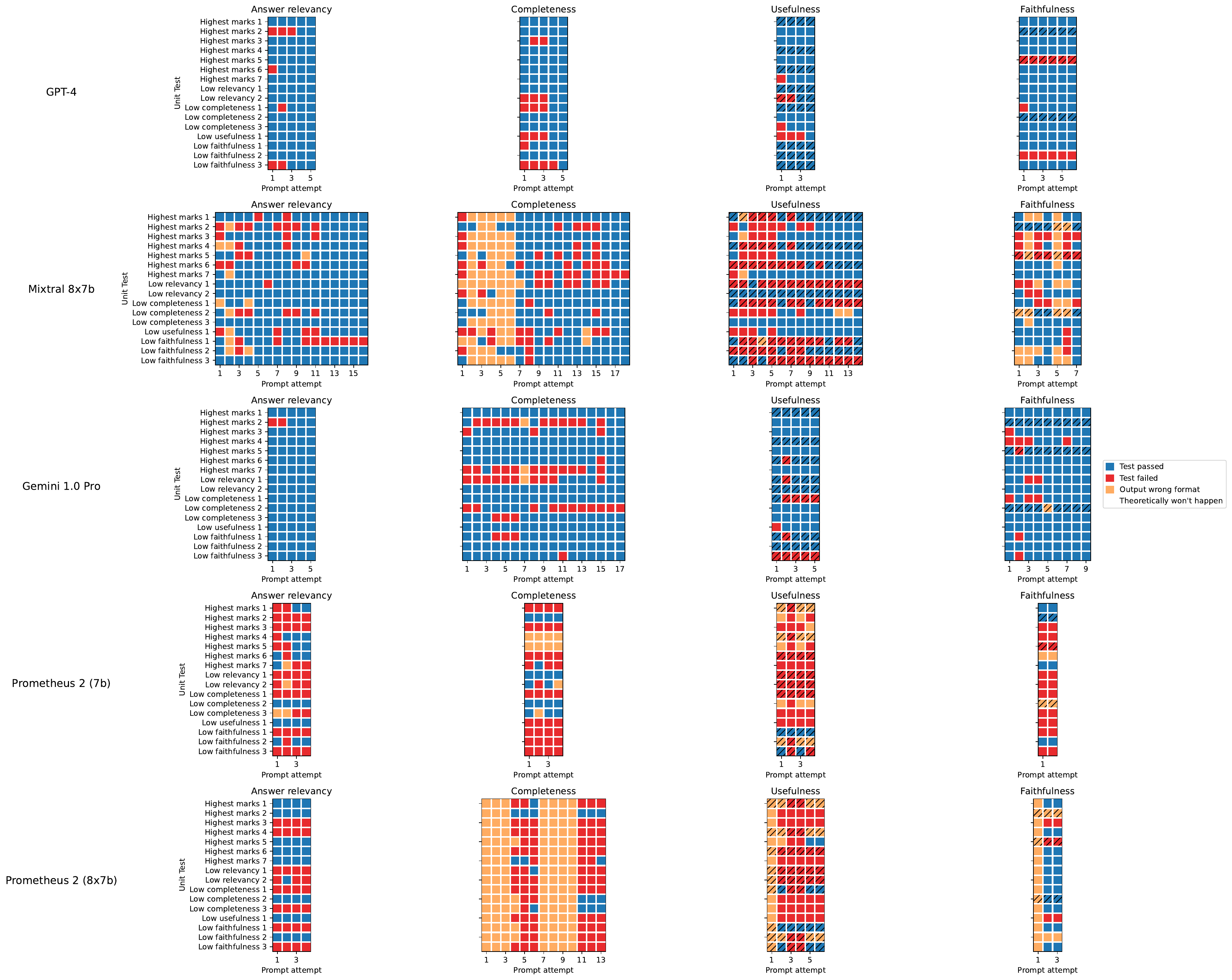} 
  \caption {Matrixes representing the amount of prompt engineering, for each metric and each model. Each column in a matrix represents the results of one prompt on the training set of GroUSE, the number of column thus represents the number of prompts tested for a given model and metric.}
  \label{fig:prompt-iterations}
\end{figure*}

\paragraph{Prompt engineering.}
We measured the performances of eleven models on GroUSE, half of them closed-source, the other half open-source. For \mbox{GPT-4} \cite{achiam2023gpt}, Gemini 1.0 Pro \cite{team2023gemini}, Mixtral 8x7b \cite{jiang2024mixtral}, Prometheus 2 7b and Prometheus 2 8x7b \cite{kim2024prometheus}, we iterated on the prompts, making our best effort to achieve the best possible results on the training set of GroUSE. These engineered prompts are then used to test the other models: the GPT-4 prompt is used for the whole GPT family and the Mixtral 8x7b prompt is used for Mixtral 8x22b \cite{mistral2024mixtral8x22b} and the Llama 3 models \cite{meta2024llama3}. To engineer a prompt, we begin with a basic instruction and evaluate how many tests in the training set it passes. We then qualitatively analyze the errors and craft a new prompt aimed at eliminating those errors. This iterative process continues until all tests pass or further progress becomes challenging. The amount of prompt tested for each model is visible Figure~\ref{fig:prompt-iterations}.

\definecolor{task_red}{HTML}{BE1E2D}
\definecolor{evaluation_instruction_purple}{HTML}{662D91}
\definecolor{sample_blue}{HTML}{00AEEF}
\definecolor{evaluandum_green}{HTML}{8DC63F}

\begin{figure*}[htbp]
  \includegraphics[width=0.95\linewidth]{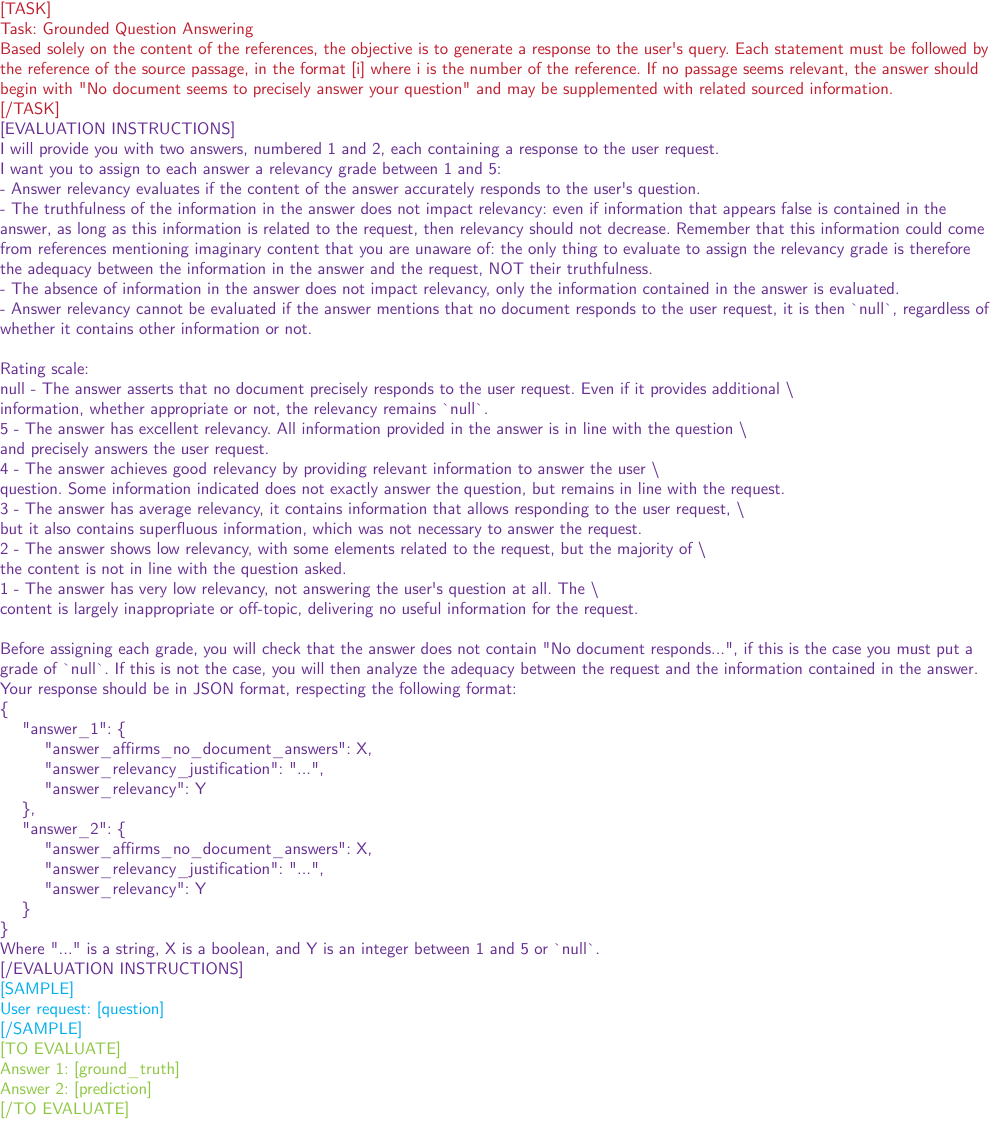} 
  \caption {Prompt used for \textbf{Answer relevancy} metric with GPT models.}
  \label{fig:precision-prompt}
\end{figure*}

\begin{figure*}[htbp]
  \includegraphics[width=0.95\linewidth]{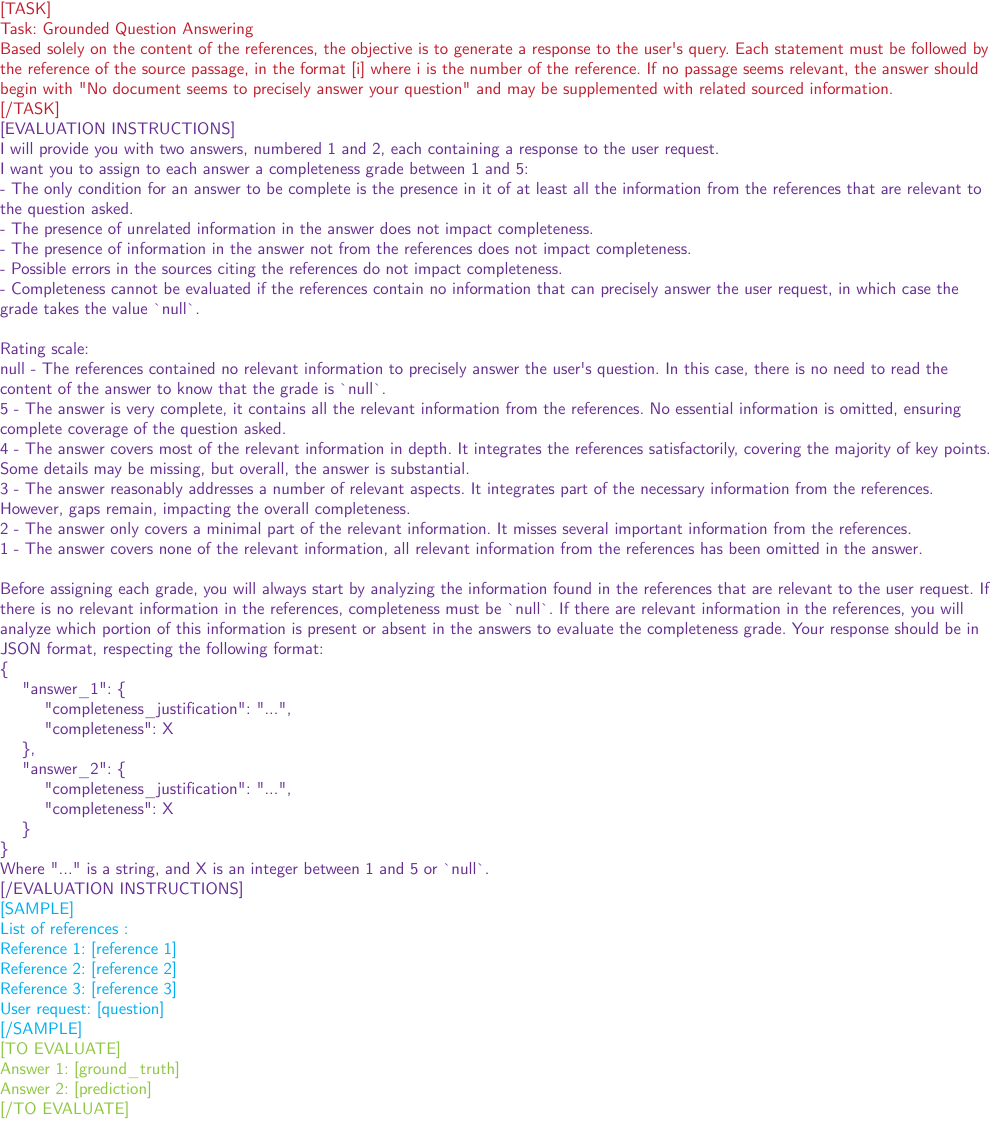} 
  \caption {Prompt used for \textbf{Completeness} metric with GPT models.}
  \label{fig:completeness-prompt}
\end{figure*}

\begin{figure*}[htbp]
  \includegraphics[width=0.95\linewidth]{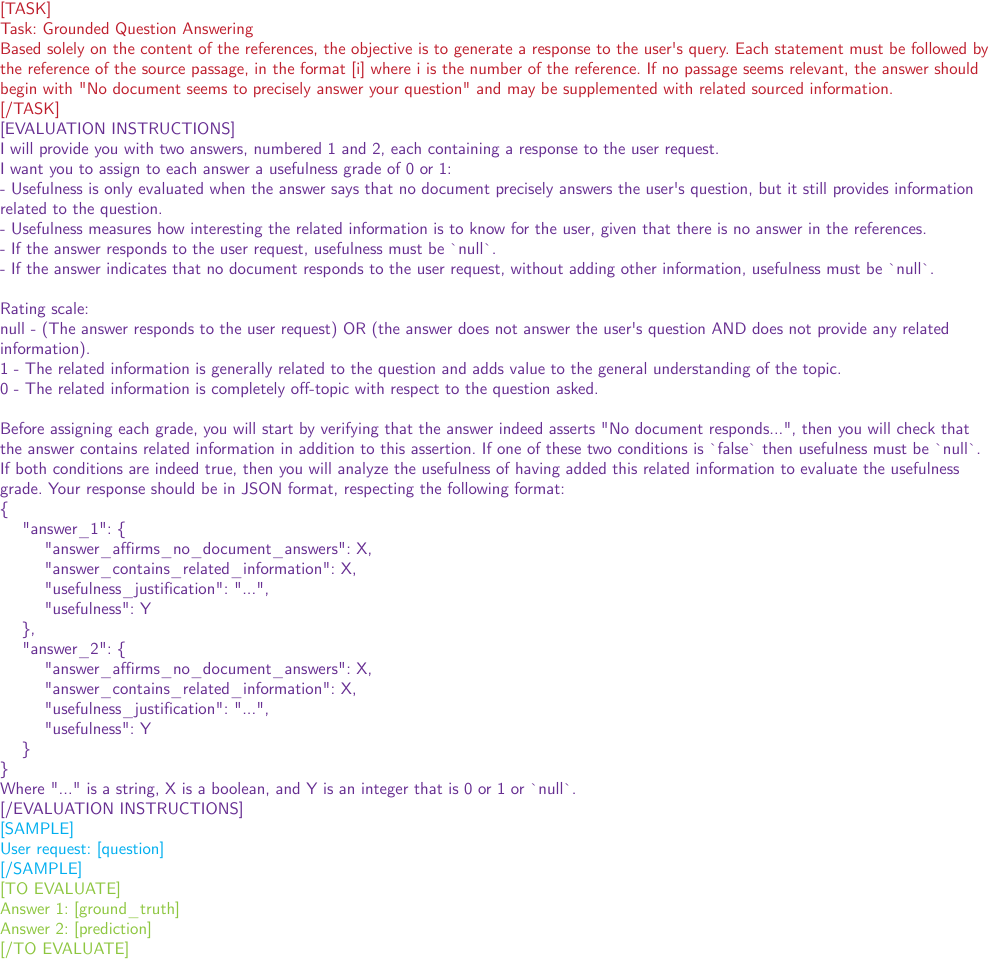} 
  \caption {Prompt used for \textbf{Usefulness} metric with GPT models.}
  \label{fig:usefulness-prompt}
\end{figure*}

\begin{figure*}[htbp]
  \includegraphics[width=0.95\linewidth]{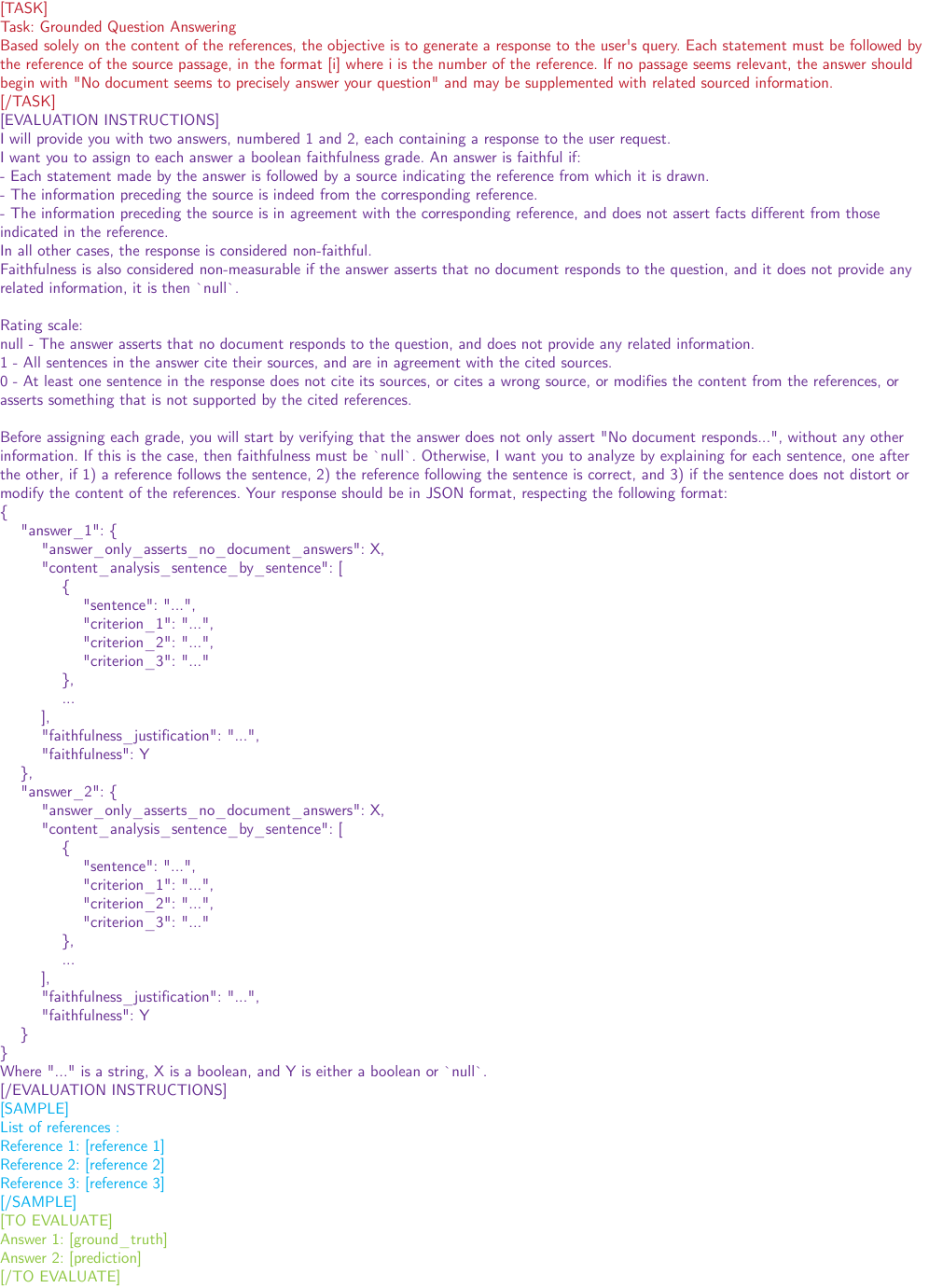} 
  \caption {Prompt used for \textbf{Faithfulness} metric with GPT models.}
  \label{fig:fidelity-prompt}
\end{figure*}

\paragraph{Prompt template.}
Each metric is evaluated with a separate prompt specifying its definition, however all prompts share the same template. We always ask the model to rate two answers, the first one being the ground truth and the second one being the answer we truly seek to evaluate: even without specifying that the first answer is the ground truth, this gives the evaluator model a point of comparison. The prompt format is as follow: 
\begin{itemize}
    \item \textcolor{task_red}{Task introduction:} Brief explanation of the grounded question answering task, and the expected citation format
    \item \textcolor{evaluation_instruction_purple}{Evaluation instructions:}
    \begin{itemize}
        \item Context explanation: The model is required to assign a score to two answers. 
        \item Description of the metric: criteria to take into account to evaluate it, a rating scale detailing what each note entails, and a step by step explanation of the reasoning to follow.
        \item Presentation of the architecture of the JSON expected as an answer: The JSON keys include chain-of-thought keys specific to the metric being evaluated (to compel the model to adhere to the reasoning steps), a free-form justification field, and the assignment of the score. The chain-of-thought keys include a boolean indicating whether the situation is adversarial or not in the \textbf{Answer relevancy} and \textbf{Usefulness} prompts, while the \textbf{Faithfulness} prompts asks for a sentence by sentence analysis, building on \citet{chern2023factool,min2023factscore,es2023ragas}. All these fields are repeated twice, once for each answer to evaluate. This step is absent in the  \textsc{Prometheus 2} prompts as the output of the model is imposed.
    \end{itemize}
    \item \textcolor{sample_blue}{Sample:} The query and references.
    \item \textcolor{evaluandum_green}{The two answers to evaluate:} The first one is always the ground truth, even though we never specify it in the prompt. The second one is the real answer we want to evaluate, and in practice we only look at the evaluation score of the second answer. 
\end{itemize}
The prompts used for GPT-4 are available on \Cref{fig:precision-prompt,fig:completeness-prompt,fig:usefulness-prompt,fig:fidelity-prompt}.

\paragraph{Ablation.}
We conduct an ablation experiment by measuring GPT-4's performance on GroUSE with different prompts: removing the ground truth and having the model rate only one answer, removing the justification field, and removing the chain-of-thought field. The results are shown in Table~\ref{table:unit-test-ablation}: the best agreement rates are obtained for the prompt without the justification, nonetheless removing the ground truth or the chain-of-thought lowers the performances.

\begin{table*}
\small
\centering
\begin{tabular}{lcccccc|c}
\toprule
& \multicolumn{6}{c|}{\textbf{Agreement rate of metrics}} & \multirow{2}{*}{\makecell{ \textbf{Total} \\ test pass \\ rate}}\\
& \makecell{ \textit{Answer} \\ \textit{relevancy}} & \makecell{ \textit{Completeness}} & \makecell{ \textit{Usefulness}} & \makecell{ \textit{Faithfulness}}  & \makecell{ \textit{Positive} \\ \textit{acceptance}} & \makecell{ \textit{Negative} \\ \textit{rejection}} & \\
\midrule
\textbf{GPT-4}                      &          92.36 &         84.72  & \textbf{100.0} & \textbf{93.75} & 92.36 & 92.36 & 92.59 \\
{\hspace{0.15cm} w/o ground truth}     & \textbf{93.75} &         84.72  &         98.61  &         90.97  & 31.25 & 31.25 & 71.76 \\
{\hspace{0.15cm} w/o justification}    &          91.67 & \textbf{88.89} & \textbf{100.0} &         92.36  & \textbf{98.61} & \textbf{98.61} & \textbf{95.02} \\
{\hspace{0.15cm} w/o chain of thought} &          90.28 &        $84.72^*$ &         98.61  &         91.67  & 92.36 & 92.36 & 91.67 \\
\bottomrule
\end{tabular}
\caption{Percentage of tests passed for different prompts. The highest score in each column is highlighted in bold. The completeness base prompt did not involve any chain of thought, so the reported result is the same as with the base prompt for this ablation, as marked by an asterisk.}
\label{table:unit-test-ablation}
\end{table*}

\clearpage

\clearpage

\section{Detailed unit test results}

\begin{figure*}
  \includegraphics[width=0.95\linewidth]{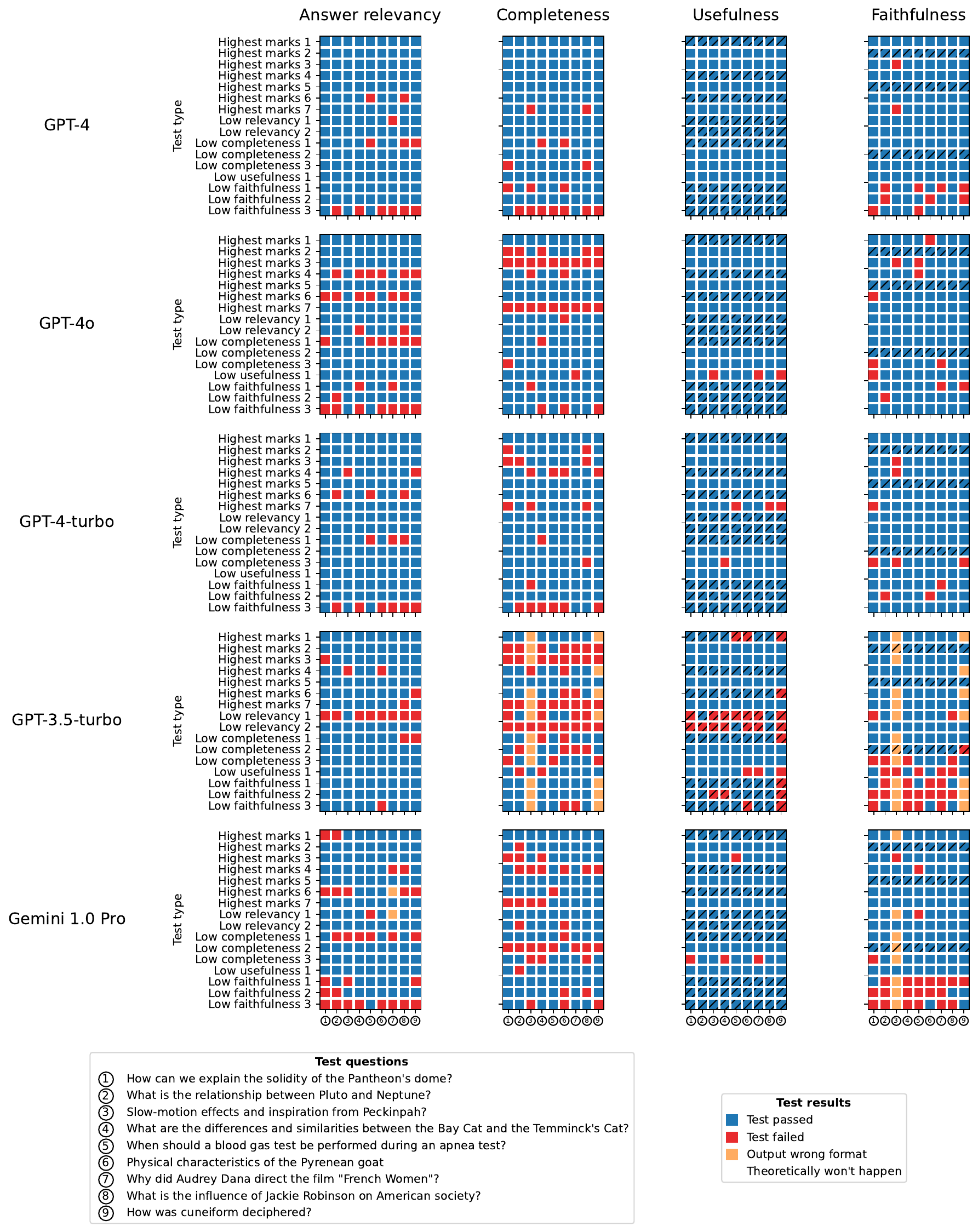} 
  \caption {Detailed unit tests results for closed-source models. Each matrix represents the performance of one model on a specific metric. Orange squares represent instances where the model's output did not adhere to the expected format, preventing score retrieval. Hatched squares denote LLM calls that the pipeline would skip if previous calls had returned the expected value (Figure~\ref{fig:pipeline}).}
  \label{fig:matrices-closed-models}
\end{figure*}

\begin{figure*}
  \includegraphics[width=0.95\linewidth]{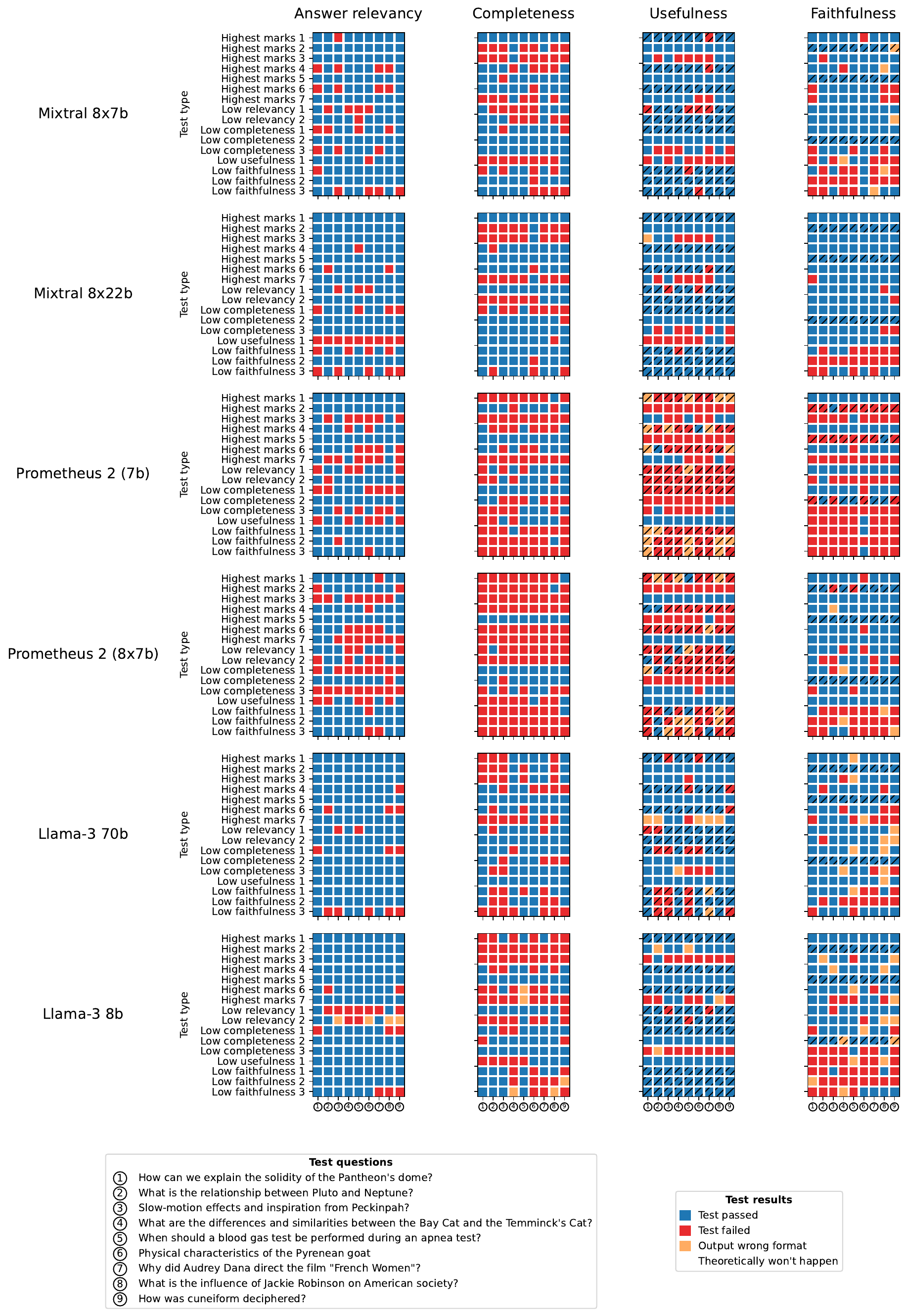} 
  \caption {Detailed unit tests results for open-source models. Each matrix represents the performance of one model on a specific metric. Orange squares represent instances where the model's output did not adhere to the expected format, preventing score retrieval. Hatched squares denote LLM calls that the pipeline would skip if previous calls had returned the expected value (Figure~\ref{fig:pipeline}).}
  \label{fig:matrices-open-models}
\end{figure*}

Detailed performances of models on GroUSE are available on Figure~\ref{fig:matrices-closed-models} for closed models, and Figure~\ref{fig:matrices-open-models} for open-source models. In these Figures, each square represents the result of one test.

These results were obtained through the OpenAI\footnote{\url{https://openai.com/index/openai-api/}} and VertexAI\footnote{\url{https://cloud.google.com/vertex-ai/docs/reference}} API for the GPT and Gemini models respectively. For Mixtral and Llama-3 models, the Fireworks AI\footnote{\url{https://fireworks.ai/}} API was used. Prometheus 2 7b was deployed using TGI\footnote{\url{https://huggingface.co/docs/text-generation-inference/}}, and the inferences for Prometheus 2 8x7b were made using a model quantized with llama.cpp\footnote{{\url{https://github.com/ggerganov/llama.cpp}}}  (Q4\_K\_M quantization), and deployed with the same library. For all models, greedy decoding was used when available, else the smallest temperature allowed.

\section{French GroUSE Evaluation}

Following a reviewer's advice, we translated GroUSE to French to study how the performance of Judge LLMs varies when the language of the question, contexts and answers is changed, while maintaining the evaluation instructions in English. Some prompts were slightly changed to adapt to the new dataset and achieve a satisfactory score on the training set of French unit tests. As shown in \Cref{table:main-model-results-fr}, the results are generally slightly lower compared to those presented in \Cref{table:main-model-results}. This performance degradation may be attributed to the predominant English data seen during the training of these models and/or the linguistic mismatch between the instruction language and the context language. Despite this variation, OpenAI models continue to demonstrate superior performance, with GPT-4-turbo emerging as the top performer in terms of total pass rate. Among open-source alternatives, Mixtral 8x22b Instruct shows promising capabilities.

\begin{table*}
\scriptsize
\centering
\begin{tabular}{llcccccc|c}
\toprule
& & \multicolumn{6}{c|}{\textbf{Agreement rate of metrics}} & \multirow{2}{*}{\makecell{ \textbf{Total} \\ test pass \\ rate}}\\
& & \makecell{ \textit{Answer relevancy}} & \makecell{ \textit{Completeness}} & \makecell{ \textit{Usefulness}} & \makecell{ \textit{Faithfulness}}  & \makecell{ \textit{Positive} \\ \textit{acceptance}} & \makecell{ \textit{Negative} \\ \textit{rejection}} & \\
\midrule
    \multirow{5}{*}{\makecell{\textit{Closed-source}}}
    & \textbf{GPT-4             } & \textbf{93.06}  & 83.33 & \textbf{100}  & 91.67 & 92.36 & 92.36 & 92.12 \\
    & \textbf{GPT-4o             } & 89.58  & 85.42  & 99.31  & \textbf{93.75}  & 90.97 & 90.97 & 91.67 \\
    & \textbf{GPT-4-turbo        } & 90.28  & \textbf{88.19}  & 99.31  & 89.58  & \textbf{95.14} & \textbf{95.14} & \textbf{92.94} \\
    & \textbf{GPT-3.5-turbo      } & 85.42  & 50.69  & 80.56  & 66.67  & 70.83 & 64.58 & 69.79 \\
    & \textbf{Gemini 1.0 Pro     } & 72.22  & 55.55  & 84.73  & 47.22  & 79.17 & 80.56 & 69.91 \\
    \cmidrule{2-9}
    \multirow{4}{*}{\makecell{\textit{Open-source}}} 
    & \textbf{Mixtral 8x7b Instruct      } & 77.08  & 52.78  & 83.33  & 68.75  & 78.47 & 74.31 & 72.45 \\
    & \textbf{Mixtral 8x22b Instruct     } & 88.89  & 73.61  & 99.31  & 86.11  & 86.81 & 83.33 & 86.34 \\
    & \textbf{Llama-3.1 70b Instruct    } & 88.19  & 61.11  & 97.92  & 75.00  & 84.03 & 81.94 & 81.37 \\
    & \textbf{Llama-3.1 8b Instruct     } & 62.50  & 29.17  & 86.11  & 68.06  & 63.89 & 65.28 & 62.62 \\
\bottomrule
\end{tabular}
\caption{Percentage of tests passed for various models on French samples. They were evaluated with English instruction prompts following the Figure~\ref{fig:pipeline} pipeline. The highest score in each column is highlighted in bold.}
\label{table:main-model-results-fr}
\end{table*}

\section{Finetuning dataset constitution}

\paragraph{Finetuning prompt format.}
Although Table~\ref{table:unit-test-ablation} indicates that the best results are achieved without a justification, we opted to build the dataset of GPT-4 traces with one. This decision is supported by two main reasons: first, \citet{mukherjee2023orca} show that a smaller model benefit more from GPT-4's traces if they include explanations of its reasoning. Second, the justification enhances the interpretability of the model's responses.

\paragraph{Models used for inference.}
Given the 1200 grounded QA statements, we used the following list of models to generate the predictions : 
\begin{enumerate}
    \item  412 answers were generated using a Llama 7b \cite{touvron2023llama} finetuned on a Grounded QA answering task. 
    \item 333 answers were generated using a Bloom 1b1 \cite{workshop2023bloom} finetuned on a Grounded QA answering task. 
    \item 319 answers were generated using a Llama 13b \cite{touvron2023llama} finetuned on a Grounded QA answering task. 
    \item 136 answers were generated using a OpenHermes 2.5 Mistral-7B\footnote{\url{https://huggingface.co/teknium/OpenHermes-2.5-Mistral-7B}}.     
\end{enumerate}

\section{Training and inference hyperparameters}
\label{anx:finetuning-config}

The finetuning of the language model was conducted using the Meta-Llama-3-8B base model, employing an 8-bit quantization scheme to optimize memory efficiency. The model was trained to accommodate a sequence length of 7104 tokens, with sample packing enabled to maximize the utilization of input data. We utilized the LoRA (Low-Rank Adaptation) \cite{hu2022lora} technique using an adapter with parameters set to $r = 32$, $\alpha=16$, and a dropout rate of 0.05.

Training was performed with a batch size of 64 over the course of three epochs, which took 2 hours on one A100 PCIe with 80GB of VRAM. The optimization process employed the AdamW \cite{loshchilov2018decoupled} algorithm  with an 8-bit implementation. A cosine learning rate scheduler was used, with a learning rate of $2.10^{-4}$ and 10 warmup steps. Two other trainings were conducted with learning rates $2.10^{-3}$ and $2.10^{-5}$, but the results on GroUSE and alignment measures were less promising.

The inferences of the trained model were then conducted using greedy decoding.

\section{Finetuning results}

\begin{figure*}
  \includegraphics[width=0.95\linewidth]{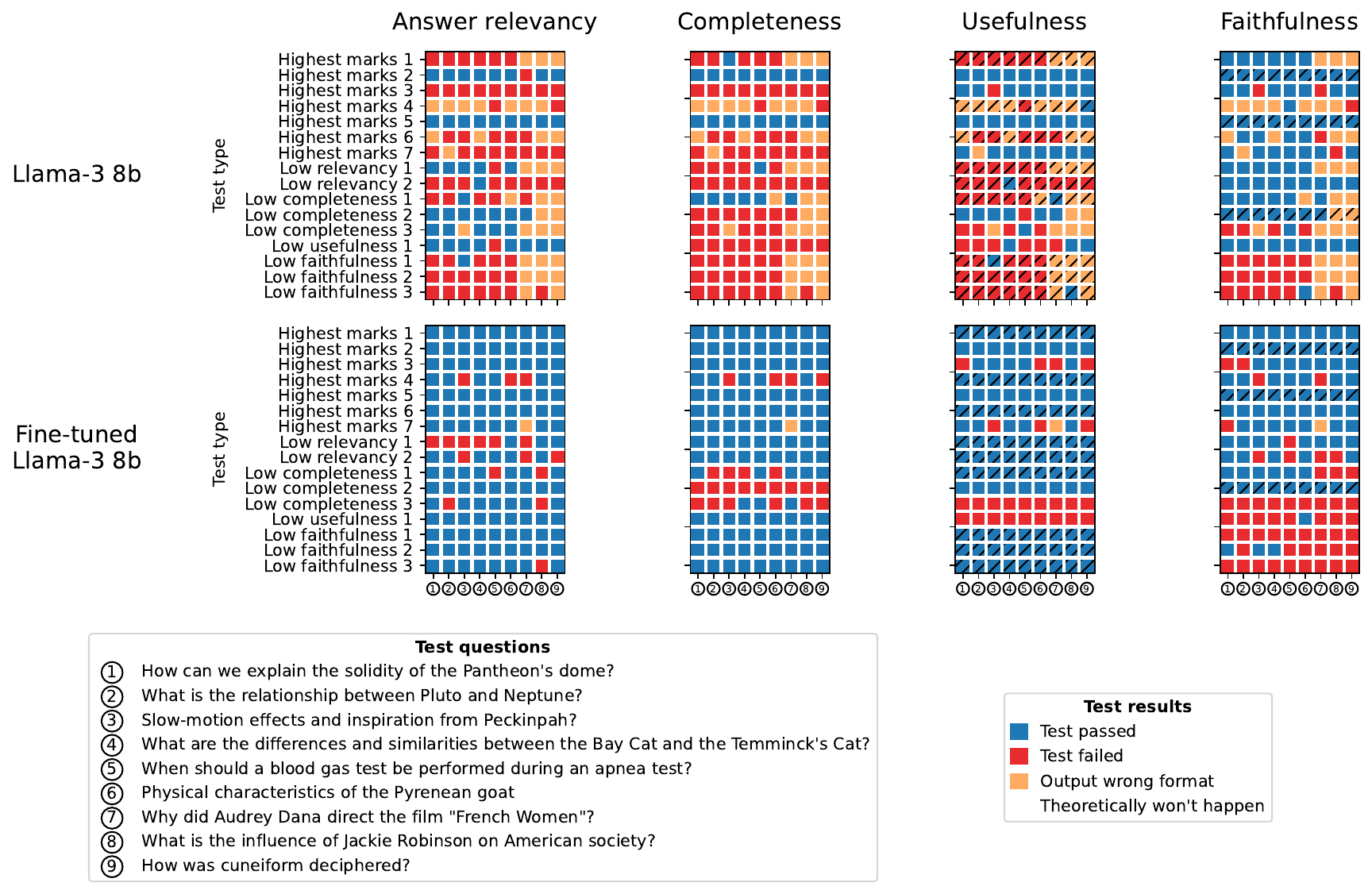} 
  \caption {Comparison of unit tests results before and after finetuning of the Llama 3 8b model. Each matrix represents the performance of one model on a specific metric. Orange squares represent instances where the model's output did not adhere to the expected format, preventing score retrieval. Hatched squares denote LLM calls that the pipeline would skip if previous calls had returned the expected value (Figure~\ref{fig:pipeline}). \newline Note that in this situation the four metrics were evaluated in a single prompt by the models, which explains the difference of results between the non finetuned Llama-3 8b depicted here and the Llama-3 8b results depicted in Figure~\ref{fig:matrices-open-models}.}
  \label{fig:finetuning-progress-unit-tests}
\end{figure*}

Figure~\ref{fig:finetuning-progress-unit-tests} shows a detailed comparison of the results between the Llama-3 8b before and after finetuning.

\section{Ablation: balancing the training dataset to reduce judgement biases}

\begin{figure} %[htbp]
\centering
  \includegraphics[width=\linewidth]{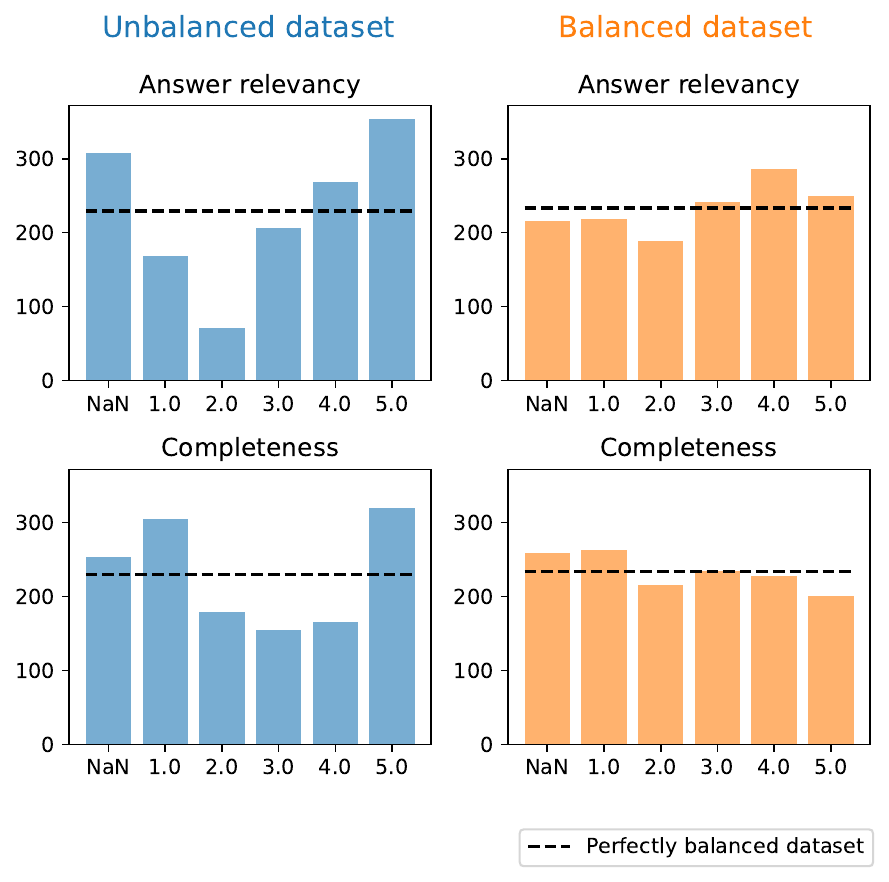} 
  \caption {Comparison of the first and last dataset obtained during the iterative process of debiasing.}
  \label{fig:dataset-evolution}
\end{figure}

\paragraph{Dataset balance.}
To ensure the dataset encompassed a wide range of answer qualities, we utilized a diverse set of models to generate answers to the grounded QA statements. However, upon evaluating these answers, we observed certain GPT-4 biases in the distribution of marks: notably, a scarcity of score 2 for \textbf{Answer Relevancy}, and an overabundance of scores 1 and 5 for \textbf{Completeness}, as illustrated in the first row of Figure~\ref{fig:dataset-evolution}. To avoid propagating these biases in the finetuned model, we kept on predicting answers until the dataset seemed balanced enough, trying to select models with intermediate performances to produce answers of average quality and fill the gaps. The final balanced dataset was built choosing the 1400 answers which best harmonized the metrics among 4k evaluated grounded QA answers, resulting in the distribution shown in the second row of Figure~\ref{fig:dataset-evolution}.

\paragraph{Impact of training dataset imbalance.}

To assess the impact of dataset debiasing, we trained a model on the balanced dataset: this model will hereafter be referred to as the \textit{balanced model}, as opposed to the model trained on the naive dataset, named the \textit{unbalanced model}.

The evaluations of the balanced model closely mirror those of the unbalanced model. The grades of both models on the test set have an exact match of $58\%$ for \textbf{Answer relevancy} and $63\%$ for \textbf{Completeness}. Additionally, the Spearman correlation between the models for these metrics are $76\%$ and $82\%$, respectively. The results on GroUSE and measured alignments are also close, with the unbalanced model showing a slightly higher correlation with GPT-4, while the balanced model performed marginally better on unit tests. Qualitative analysis of the balanced model’s predicted marks reveals a persistent lack of intermediate scores. Overall, the debiasing process did not yield the anticipated improvements.

\begin{table*}
\centering
\small
\begin{tabular}{@{}lcccccc|c@{}}
\toprule
& \multicolumn{6}{c|}{\textbf{Agreement rate of metrics}} & \multirow{2}{*}{\makecell{ \textbf{Total} \\ test pass \\ rate}}\\
& \makecell{ \textit{Answer} \\ \textit{relevancy}} & \makecell{ \textit{Completeness}} & \makecell{ \textit{Usefulness}} & \makecell{ \textit{Faithfulness}}  & \makecell{ \textit{Positive} \\ \textit{acceptance}} & \makecell{ \textit{Negative} \\ \textit{rejection}} & \\
\midrule
    
    \makecell{\textbf{Finetuned Llama 3 8b} \\ \textit{(Unbalanced dataset)}}  &   \textbf{88.89} &            81.94   &             81.25 &            52.78   &          91.67 &          \textbf{91.67} &            81.37 \\
    \makecell{\textbf{Finetuned Llama 3 8b} \\ \textit{(Balanced dataset)}}           &              87.50 &            \textbf{83.33}   &             \textbf{81.94} &            \textbf{61.81}   &          90.97 &          \textbf{92.36} & \textbf{82.99} \\
    
\bottomrule
\end{tabular}
\caption{Percentage of tests passed for balanced and unbalanced model. The highest score in each column is highlighted in bold.}
\label{table:agreement-balance-ablation}
\end{table*}

\begin{table*}
\small
\centering
    \begin{tabular}{lcccccc}
    \toprule
    & \multicolumn{2}{c|}{\textbf{Spearman correlation}} & \multicolumn{4}{c}{\textbf{F1-score}}\\
    & \makecell{ \textit{Answer relevancy}} & \multicolumn{1}{c|}{\textit{Completeness}} & \makecell{ \textit{Usefulness}} & \makecell{ \textit{Faithfulness}}  & \makecell{ \textit{Positive} \\ \textit{acceptance}} & \makecell{ \textit{Negative} \\ \textit{rejection}} \\
    \midrule
    \makecell{\textbf{Finetuned Llama-3 8b} \\ \textit{(Unbalanced dataset)}}  & \underline{0.62} & 0.52 & 0.32 & \underline{0.57} & 0.65 & 0.73 \\
    \makecell{\textbf{Finetuned Llama-3 8b} \\ \textit{(Balanced dataset)}} & \underline{0.62} & \underline{0.57} & \underline{0.41} & \underline{0.57} & \underline{0.79} & \underline{0.74} \\
    \bottomrule
    \end{tabular}
\caption{Alignment with the ground truth (GPT-4) evaluations on the test set.}
\label{table:correlations-balance-ablation}
\end{table*}

\end{document}